\documentclass[11pt,a4paper]{report}
\usepackage[utf8]{inputenc}
\usepackage{amsmath}
\usepackage{amsfonts}
\usepackage{amssymb}
\usepackage{color}
\usepackage{hyperref}
\usepackage{adjustbox}
\usepackage{float}
\usepackage{subcaption}
\usepackage{array}
\usepackage{eurosym}
\usepackage[some,top]{background}
\usepackage{fancyhdr}
\usepackage{transparent}
\usepackage{indentfirst}
\usepackage{etoolbox}
\usepackage{pdfpages}
\usepackage{longtable}

\usepackage[a4paper, total={15.4cm, 24cm}]{geometry}
\makeatletter
\patchcmd{\@makechapterhead}{\vspace*{50\p@}}{}{}{}%
\patchcmd{\@makeschapterhead}{\vspace*{50\p@}}{}{}{}%
\patchcmd{\@makechapterhead}{\vskip 40\p@}{\vskip 10\p@}{}{}%
\patchcmd{\@makeschapterhead}{\vskip 40\p@}{\vskip 10\p@}{}{}%
\patchcmd{\@makechapterhead}{\vskip 20\p@}{\vskip 10\p@}{}{}%
\patchcmd{\@makeschapterhead}{\vskip 20\p@}{\vskip 10\p@}{}{}%
\makeatother

\usepackage{graphicx} % LaTeX
\DeclareGraphicsExtensions{.bmp,.png,.pdf,.jpg}
\author{Issey Masuda Mora}
\title{Open-Ended Visual Question-Answering}
\newsavebox{\mygraphic}
\sbox{\mygraphic}{\transparent{0.7}\includegraphics[keepaspectratio,
 width=0.15\textwidth]{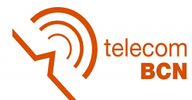}} 

\newsavebox{\mygraphicupc}
\sbox{\mygraphicupc}{\transparent{0.7}\includegraphics[keepaspectratio, width=0.25\textwidth]{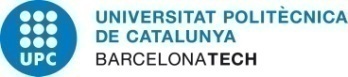}} 

\fancypagestyle{plain}{
\fancyhead{}

\fancyhead[R]{
\begin{picture}(0,0)
\put(400,0){\usebox{\mygraphic}}
\end{picture}} 
\fancyhead[L]{
\begin{picture}(0,0)
\put(0,0){\usebox{\mygraphicupc}}
\end{picture}}
}

\begin{document}

\begin{titlepage}

	\begin{center}
		% University logo
		\begin{figure}[htb]
		\centering
			\includegraphics[width=1\textwidth]{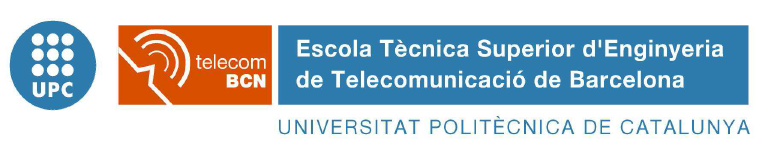} 
		\end{figure}

		\vspace{1cm}
        
		% Title
		\textbf{ \Huge Open-Ended Visual Question-Answering}\\
%		\rule{\textwidth}{0.5pt}
%		\vspace{-2ex}
%		\centering
		%\vspace{1cm}
		\vspace*{0.3in}
		%\rule{120mm}{0.05mm}\\
		%\rule{\textwidth}{0.03mm}\\
		\vspace*{0.1in}
		%subtitle
		{\Large  
        A Degree Thesis \\
        Submitted to the Faculty of the \\
        Escola Tècnica Superior d'Enginyeria de Telecomunicació de Barcelona}\\
        \vspace{1cm}
        {\Large 
        In partial fulfilment \\
        of the requirements for the degree in \\
        SCIENCE AND TELECOMMUNICATION TECHNOLOGIES ENGINEERING}\\
		\vspace{5cm}
		\begin{tabular}{ll}
		
			% Student name
			\large{\textbf{Author:}}	&
			\large{Issey Masuda Mora} \\

			% Advisors' name
			\large{\textbf{Advisors:}}	&
			\large{Xavier Giró i Nieto, Santiago Pascual de la Puente} \\	
			\vspace{1cm}	
		\end{tabular}
        
		%Universities
		\large{\textbf{Universitat Politècnica de Catalunya (UPC)}}\\
		\large{\textbf{June 2016}}\\
	\end{center}
\end{titlepage}

\newpage\null\thispagestyle{empty}\newpage

\chapter*{}
\pagenumbering{gobble}

\vspace*{8cm}
\begin{flushright}
\Large \emph{To my grandfather}
\end{flushright}
\chapter*{Abstract}
\addcontentsline{toc}{chapter}{Abstract}
\pagenumbering{arabic}

This thesis studies methods to solve Visual Question-Answering (VQA) tasks with a Deep Learning framework.

As a preliminary step, we explore Long Short-Term Memory (LSTM) networks used in Natural Language Processing (NLP) to tackle Question-Answering (text based).

We then modify the previous model to accept an image as an input in addition to the question. For this purpose, we explore the VGG-16 and K-CNN convolutional neural networks to extract visual features from the image. These are merged with the word embedding or with a sentence embedding of the question to predict the answer.

This work was successfully submitted to the Visual Question Answering Challenge 2016, where it achieved a 53,62\% of accuracy in the test dataset.

The developed software has followed the best programming practices and Python code style, providing a consistent baseline in Keras for different configurations. The source code and models are publicly available at \url{https://github.com/imatge-upc/vqa-2016-cvprw}.
\chapter*{Resum}
\addcontentsline{toc}{chapter}{Resum}

Aquesta tesis estudia mètodes per resoldre tasques de \emph{Visual Question-Answering} emprant tècniques de \emph{Deep Learning}.

Com a pas preliminar, explorem les xarxes \emph{Long Short-Term Memory} (LSTM) que s'utilitzen en el Processat del Llenguatge Natural (NLP) per atacar tasques de \emph{Question-Answering} basades únicament en text.

A continuació modifiquem el model anterior per acceptar una imatge com a entrada juntament amb la pregunta. Per aquest propòsit, estudiem l'ús de les xarxes convolucionals VGG-16 i K-CNN per tal d'extreure els descriptors visuals de la imatge. Aquests descriptors són fusionats amb el \emph{word embedding} o \emph{sentence embedding} de la pregunta per poder predir la resposta.

Aquest treball ha estat presentat al \emph{Visual Question Answering Challenge 2016}, on ha obtingut una precisió del 53,62\% en les dades de test.

El software desenvolupat ha emprat bones pràctiques en programació i ha seguit les directrius d'estil de Python, proveïnt un projecte base en Keras consistent a diferents configuracions. El codi font i els models són públics a \url{https://github.com/imatge-upc/vqa-2016-cvprw}.

\chapter*{Resumen}
\addcontentsline{toc}{chapter}{Resumen}

Esta tesis estudia métodos para resolver tareas de \emph{Visual Question-Answering} usando técnicas de \emph{Deep Learning}.

Como primer paso, exploramos las redes \emph{Long Short-Term Memory} (LST) que se usan en el Procesado del Lenguaje Natural (NLP) para atacar tareas de \emph{Question-Answering} basadas únicamente en texto.

A continuación modificamos el modelo anterior para aceptar una imagen como entrada junto con la pregunta. Para este propósito, estudiamos el uso de las redes convolucionales VGG-16 y K-CNN para extraer los descriptores visuales de la imagen. Estos descriptores son fusionados con el \emph{word embedding} o \emph{sentence embedding} de la pregunta para poder predecir la respuesta.

Este trabajo se ha presentado al \emph{Visual Question Answering Challenge 2016}, donde ha obtenido una precisión del 53,62\% en los datos de test.

El software desarrollado ha usado buenas prácticas de programación y ha seguido las directrices de estilo de Python, proveyendo un proyecto base en Keras consistente a distintas configuraciones. El código fuente y los modelos son públicos en \url{https://github.com/imatge-upc/vqa-2016-cvprw}.
\chapter*{Acknowledgements}
\addcontentsline{toc}{chapter}{Acknowledgements}

I would like to thank to my tutor Xavier Giró i Nieto for his help during the whole project, for letting me join his research group and initiating me in this amazing field and for his patience in my moments of stubbornness.

I also want to thank Santiago Pascual de la Puente for the countless times he helped me throughout the project with his wise advises and knowledge in Deep Learning. 

My partners in the X-theses group deserve also a mention here as talking with them week after week about the project and listening what they have been researching has enriched this project. Together with them I would also want to thank Albert Gil for his help and support regarding the GPI cluster usage.

I would also like to thank to Marc Bolaños, Petia Radeva and the rest of the Computer Vision group in Universitat de Barcelona for their advice and for providing us with very useful data for our experiments.

Last but not least, I want to thank to my family and friends for being there when needed.

\chapter*{Revision history and approval record}
\addcontentsline{toc}{chapter}{Revision history and approval record}

\begin{table}[h]
\centering
\begin{tabular}{| m{5em}| m{10em} | m{15em} |} 
\hline
\textbf{Revision} & \textbf{Date} & \textbf{Purpose}\\ [0.5ex]
\hline
0 &  11/06/2016 &  Document creation\\ [0.5ex]
\hline
1 &  20/06/2016 &  Document finalization\\ [0.5ex]
\hline
2 &  24/06/2015 &  Document revision \& corrections\\ [0.5ex]
\hline
3 &  25/06/2015 &  Document revision \& corrections\\ [0.5ex]
\hline
4 &  26/06/2015 &  Document revision \& corrections\\ [0.5ex]
\hline
5 &  27/06/2015 &  Document revision \& corrections\\ [0.5ex]
\hline
\end{tabular}
\end{table}

\vspace{2cm}

DOCUMENT DISTRIBUTION LIST

\begin{table}[h]
\centering
\begin{tabular}{| m{15em}| m{15em}|} 
\hline
\textbf{Name} & \textbf{e-mail} \\ [0.5ex]
\hline
Issey Masuda Mora & issey.masuda@alu-etsetb.upc.edu\\ [0.5ex]
\hline
Xavier Giró i Nieto &  xavier.giro@upc.edu\\ [0.5ex]
\hline
Santiago Pascual de la Puente &  santiago.pascual@tsc.upc.edu\\ [0.5ex]
\hline
\end{tabular}
\end{table}

\vspace{2cm}

\begin{table}[h]
\centering
\begin{tabular}{| m{4em}| m{7em}| m{4em}| m{7em}| m{4em}| m{7em}|} 
\hline
\multicolumn{2}{|c|}{\textbf{Written by:}} & \multicolumn{2}{|c|}{\textbf{Reviewed and approved by:}} & \multicolumn{2}{|c|}{\textbf{Reviewed and approved by:}} \\ [0.5ex]
\hline
\textbf{Date} & 20/06/2016 & \textbf{Date} & 27/06/2016 & \textbf{Date} & 27/06/2016\\ [0.5ex]
\hline
\textbf{Name} & Issey Masuda Mora & \textbf{Name} & Xavier Giró i Nieto & \textbf{Name} & Santiago Pascual de la Puente \\ [0.5ex]
\hline
\textbf{Position} & Project Author & \textbf{Position} & Project Supervisor & \textbf{Position} & Project Co-supervisor \\ [0.5ex]
\hline
\end{tabular}
\end{table}

%%%%%%%%%%%%%%%%%%%%%%%%%%%%%%%%%%%%%%%%%%%%%%%%%%%%%%%%%%%%%%%
%%%INDEX    %%%%%%%%%%%%%%%%%%%%%%%%%%%%%%%%%%%%%%%%%%%%%%%%%%%
\tableofcontents

%%%%%%%%%%%%%%%%%%%%%%%%%%%%%%%%%%%%%%%%%%%%%%%%%%%%%%%%%%%%%%%
%%%LIST OF FIGURES %%%%%%%%%%%%%%%%%%%%%%%%%%%%%%%%%%%%%%%%%%%%
\clearpage
\addcontentsline{toc}{chapter}{\listfigurename}
\listoffigures

%%%%%%%%%%%%%%%%%%%%%%%%%%%%%%%%%%%%%%%%%%%%%%%%%%%%%%%%%%%%%%%
%%%LIST OF TABLES%%%%%%%%%%%%%%%%%%%%%%%%%%%%%%%%%%%%%%%%%%%%%%
\clearpage
\addcontentsline{toc}{chapter}{\listtablename}
\listoftables

%%%CHAPTER 1%%%%%%%%%%%%%%%%%%%%%%%%%%%%%%%%%%%%%%%%%%%%%%
\chapter{Introduction}

\section{Statement of purpose}

In the last few years the number of published papers and job offers related to Deep Learning have exploded. Both the academic world and the industry are pushing forward to speed up the developments and the research in this area. The reason is that Deep Learning has shown a great performance solving a lot of problems that were previously tackled by more classic Machine Learning algorithms and it has also opened the door to more complex tasks that we could not solve before. 

We humans are constantly asking questions and answering them. This is the way we learn, we transfer knowledge and lastly we communicate with each other. This basic framework of communication has inspired other ways of communications such as the HTTP protocol which is basically a combination of a request (question) and a response (answer). Frequently Asked Questions (FAQ) also uses this format. 

But what about machines? Artificial Intelligence is a huge science-fiction topic and it is recurrently all over the news and media, but the reality is not that far from there. Deep Neural Networks are nowadays used in our everyday life when we surf the net, when we use recommendation systems or automatic translation systems. This has also been extended to tackle Question-Answering tasks from the Natural Language Processing perspective (\emph{e.g.} Facebook AI Research presented a set of tasks, called bAbI \cite{weston2015towards}, to evaluate AI models' text understanding and reasoning).

\begin{figure}
\centering
\includegraphics[width=1\textwidth]{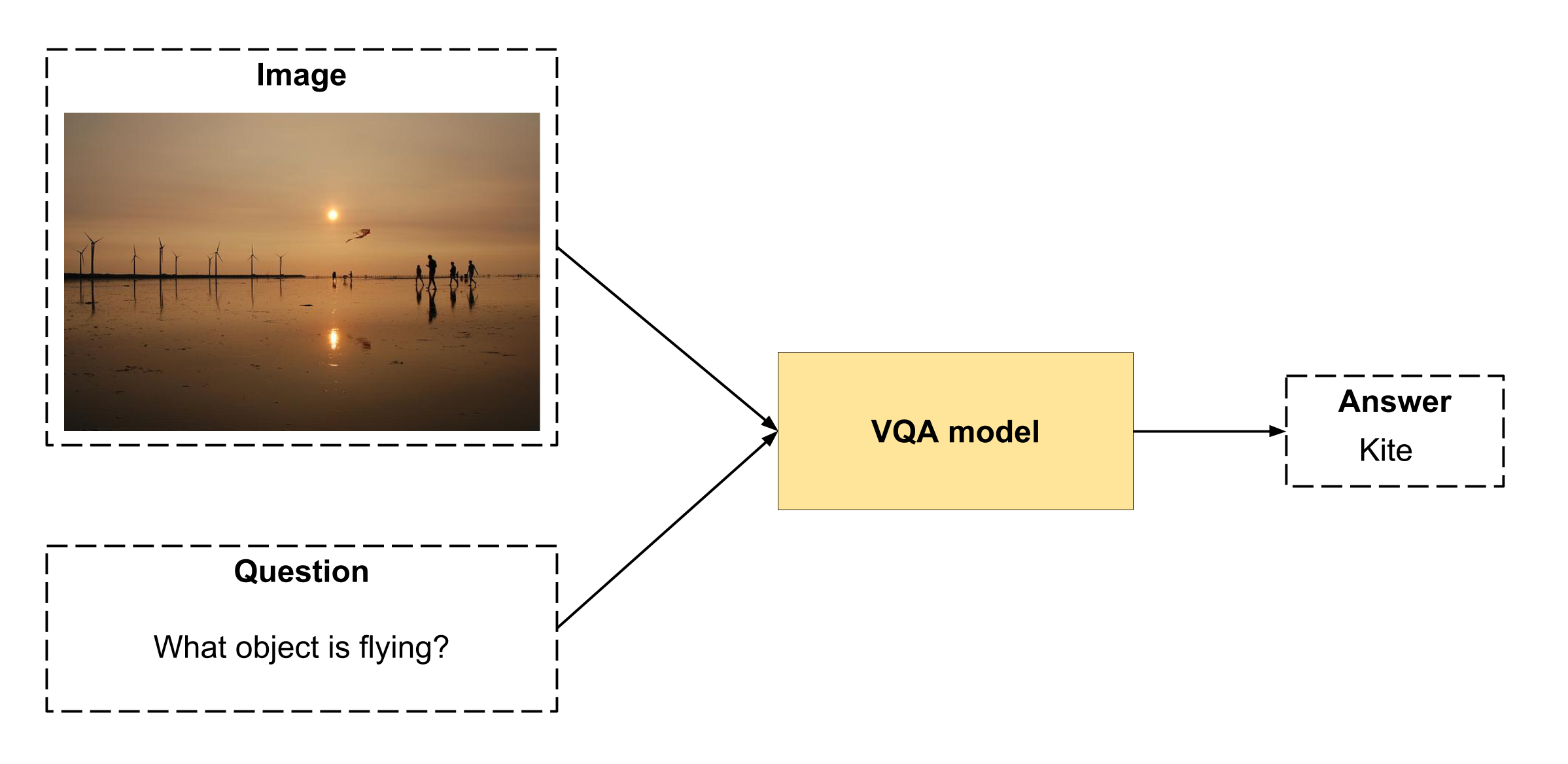}
\caption{Real example of the Visual Question-Answering dataset. The complexity of the task and the required abilities are appreciable in this example where, in order to succeed, the model needs to solve an object retrieval-like task but with the addition of having to understand the scene and the mentions of the question, \emph{e.g.} the relationship between the word 'flying' and the object position}
\label{fig:vqa-example}
\end{figure}

Visual Question-Answering has emerged as an evolution of these text-based QA systems. These models aim to be able to answer a given natural question related to a given image. One of the interests in such models is that in order to succeed in these visual QA tasks (or even just text-based QA), they need to have a much deeper level of \emph{reasoning} and understanding than other similar models, for example image captioning models. An example of a VQA task is shown in Figure \ref{fig:vqa-example}

This thesis studies new models to tackle VQA problems. The common point of all these models is that they use Convolutional Neural Networks (CNN) to process the image and extract visual features, which are a summarized representation of the image, and Long Short-Term Memory networks (LSTM), a flavor of Recurrent Neural Network (RNN), to process the question sequence. 

Based on the given context, the main objectives of this project are:
\begin{itemize}
\item Explore the techniques used for text-based Question-Answering
\item Build a model able to perform visual question-answering
\item Compare which approach is better to process the question: word embedding or sentence embedding. These are two different techniques to project a text into a space with semantic relations, meaning that you can perform some arithmetic operations with those embedding and the result will have semantic sense
\item Try different architectures and parameters to increase model's accuracy
\item Develop a reusable software project using programming good practices
\item Participate in Visual Question Answering Challenge\footnote{\url{http://www.visualqa.org/challenge.html}}, hosted as workshop in the IEEE Conference on Computer Vision and Pattern Recognition (CVPR) 2016.
\end{itemize}

Regarding the last item, we presented our results to the challenge with an accuracy of 53,62\% (details on the results chapter \ref{cha:results} and the model employed will be discussed further on the methodology chapter \ref{cha:methodology}).

We also achieved an additional goal, which was not planned at the beginning of the project. We submitted an extended abstract (you can find it in the appendices \ref{cha:appendices}) to the CVPR16 VQA workshop\footnote{\url{http://www.visualqa.org/workshop.html}} and it was accepted by the organizers. Due to this fact, we were invited to present our extended abstract and a poster in the workshop.

%%%%%%%%%%%%%%%%%%%%%%%%%%%%%%%%%%%%%%%%%%%%%%%%%%%%%%%%%%%%

\section{Requirements and specifications}

One of the main blocks of this project is the software developed to participate in the challenge and to be able to create and test different models. 

Regarding with this, the requirements of this project are the following:

\begin{itemize}
\item Develop a software that can be used in the future to keep doing research in this field, having a skeleton/base project to start with

\item Build a deep neural network model that uses NLP and CV techniques to process the question and the image respectively

\item Try different model configurations to increase the accuracy of the original model

\item Submit results to the CVPR16 VQA Challenge
\end{itemize}

The specifications are the following:

\begin{itemize}
\item Use Python as a programming language

\item Build the project using a deep learning framework. Keras\footnote{\url{http://keras.io/}} has been chosen as the framework and it can run upon Theano\footnote{\url{http://deeplearning.net/software/theano/}} or TensorFlow\footnote{\url{https://www.tensorflow.org/}} backends.
\end{itemize}

%%%%%%%%%%%%%%%%%%%%%%%%%%%%%%%%%%%%%%%%%%%%%%%%%%%%%%%%%%%%

\section{Methods and procedures}
\label{sec:methods-procedures}

This thesis represents the first attempt at solving the Visual Question Answering problem by GPI and TALP research groups at the Universitat Politècnica de Catalunya. We started to develop the project from scratch (in terms of the project itself) but using a deep learning framework called Keras.
Keras is a neural network library for Python that has been build to be easy to use and allow fast prototyping. It accomplished this by building a wrapper around another deep learning python library that is in charge of managing tensors and the low level computations. This second library, that works as a \emph{backend}, can be either Theano or TensorFlow. We have run our experiments using Keras over Theano.

Apart from these libraries, the only resource developed by other authors are the visual features of our last model. The Computer Vision group at Universitat de Barcelona provided us with the precomputed visual features of the images from the VQA dataset. They extracted these features using a special kind of CNNs called Kernelized CNN (KCNN) as proposed by Liu \cite{liu2015kernelized}. 
The KCNN method aims to provide a better vectorized representation of images than vanilla CNNs as they have a lack when the image has complex content. This model uses CNNs to extract features and then aggreagate them into a vectorial representation using the Fisher vector model.

%%%%%%%%%%%%%%%%%%%%%%%%%%%%%%%%%%%%%%%%%%%%%%%%%%%%%%%%

\section{Work Plan}
\label{sec:workplan}

The project was developed as a collaboration between the GPI and TALP research groups of the Universitat Politècnica de Catalunya. Discussions and decisions about the project were held in a regular weekly meeting, which was complemented with a second research seminar of two hours per week with other students developing their bachelor, master or Phd thesis at GPI.

The following is the workplan of this project and its deviations from the original plan. These deviations are explained in detail in the Incidents and Modifications subsection \ref{sec:modification}.

\subsection{Work Packages}
\begin{itemize}
\item WP 1: Project proposal and work plan.
\item WP 2: Introduction to deep learning and Python
\item WP 3: Introduction to (visual) question-answering tasks and Keras framework
\item WP 4: First VQA model
\item WP 5: Critical Review of the project
\item WP 6: Participate in the CVPR16 VQA Challenge
\item WP 7: Final report of the project
\item WP 8: Presentation and oral defense
\end{itemize}

\subsection{Gantt Diagram}
\begin{figure}[H]
\centering
\includegraphics[width=1\textwidth]{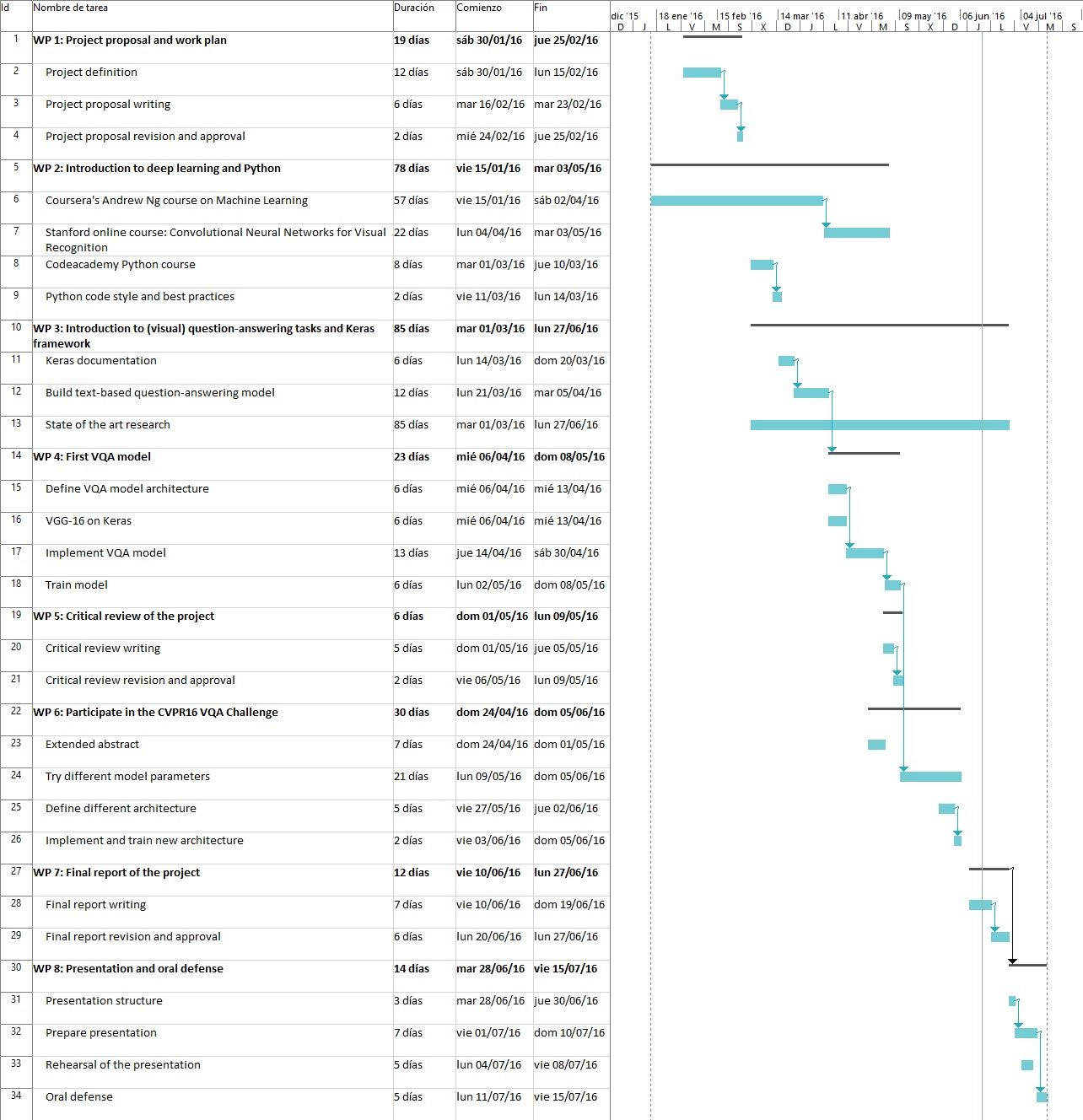}
\caption{Gantt Diagram of the project}
\label{fig:gantt}
\end{figure}

%%%%%%%%%%%%%%%%%%%%%%%%%%%%%%%%%%%%%%%%%%%%%%%%%%%%%%%%%%%%%%%%%%%%

\section{Incidents and Modifications}
\label{sec:modification}

During the project we needed to modify some work packages definition and tasks as we wanted to focus more on the Visual Question-Answering Challenge.

Initially, the goal of the project was developing a system capable of generating both questions and answers from an image. This would have medical application in patients with mild cognitive impairment (early stages of Alzheimer), who may receive an automatized reminiscence therapy based on the images captured by egocentric cameras. However, solving the VQA challenge was more feasible in terms of annotated datasets, metrics and potential impact, so it was decided to address this task first. The described medical applications are planned to be explored by other students during Fall 2016.

We also included the new task of writing an extended abstract for the VQA workshop. We decided to write and submit the extended abstract as this would gave me some expertise on paper composing and this way we could share with the community some of our ideas.

%%%CHAPTER 2%%%%%%%%%%%%%%%%%%%%%%%%%%%%%%%%%%%%%%%%%%%%%%
\chapter{State of the art}

In the past years, multidisciplinary problems of vision, language and reasoning have emerged as a trend in Artificial Intelligence (AI) research. This tasks join Computer Vision (CV), Natural Language Processing (NLP) and Knowledge Representation and Reasoning (KR) to be able to build models that can interact with both image and language input/output. However, this models still fail achieving accuracies close to human level.

Visual Question-Answering has appeared as a problem where models need to be able to perform different sub-problems of the above three fields in order to succeed. To solve this problems the models need a much deeper understanding and comprehension of the scene in the image, what the question is referring to and how the items are related.

We will revise some of the literature involved in the process of building a VQA model, from image and text processing, to the state-of-the-art approaches for VQA tasks.

%%%%%%%%%%%%%%%%%%%%%%%%%%%%%%%%%%%%%%%%%%%%%%%%%%%%%%%%%%%%%%%%%%%%%

\section{Image processing}
Deep Convolutional Neural Networks (CNN) have been proved to achieve state-of-the-art results in typical Computer Vision tasks such as image retrieval, object detection and object recognition. 

A common approach when dealing with images is to use an off-the-shelf model (VGG \cite{simonyan2014very}, AlexNet \cite{krizhevsky2012imagenet}, GoogLeNet \cite{szegedy2015going}, etc.) pre-trained to do such tasks with some large image dataset such as ImageNet\footnote{\url{http://www.image-net.org/}} \cite{deng2009imagenet} and use some of the inner-layer's outputs as a representation of the visual features of the image.

\begin{figure}[H]
\centering
\includegraphics[width=1\textwidth]{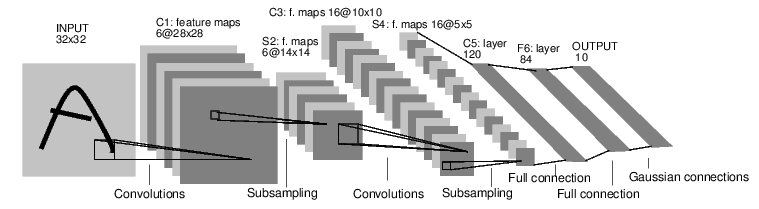}
\caption{LeNet, an example of Convolutional Neutal Network}
\label{fig:lenet}
\end{figure}

Typically these models have different types of layers, amongst the most common convolutional layers (that give the name to the CNNs) and fully-connected layers. \\
The convolutional layers used in image processing perform 2D convolutions of the previous layer output (which can be an image) where the weights specify the convolution filter. \\
In contrast, fully-connected layers take each output from the previous layer and connect them to all of its neurons, losing the spatial information so they can be seen as one dimensional. One of the most common fully-connected layers is the so called softmax layer, which is a regular fully-connected with the softmax as activation function. Its output follows a distribution-like shape, taking values from 0 to 1 and being the addition of all of them equal to 1.

%%%%%%%%%%%%%%%%%%%%%%%%%%%%%%%%%%%%%%%%%%%%%%%%%%%%%%%%%%%%%%%%%%%%%

\section{Text processing}
In order to process sequences of text, different approaches are used. For the sake of the simplicity, we will only review two of them that are important for this work.

The first one is the word embedding representation using Skip-gram technique presented by Mikolov \emph{et. al.} \cite{mikolov2013efficient}\cite{mikolov2013distributed}. This method is used to learn high-quality word vector representations. The input is usually the index of the word in a dictionary (\emph{i.e.} its one-hot code), a vector as large as the size of the vocabulary which is zero-valued except at the index position corresponding to the word. These vectors are high-dimensional (as the dictionary size can have thousands or hundred of thousands of words) and sparse due to the nature of the one-hot representation. The word embedding projects this vector into a semantic space where each word is represented by a dense vector with less dimensions. 
This technique captures semantic and syntactic relationships between words and also encodes many linguistic patterns based on the context where the words appear. These patterns can be expressed as algebraic operations, \emph{e.g.} embed("King") - embed("Man") + embed("Woman") has as the closest vector the embedding of "Queen".

The logical evolution of this representation is what is called sentence embedding. Word embedding fails at capturing the long-term dependencies between words that appear together in a sentence. To solve this problem, sentence embedding uses Recurrent Neural Networks (RNN) with Long Short-Term Memory cells (LSTM) to increasingly accumulate richer information about the sentence. \\
Note that RNN are deep neural networks with a memory state that allows them to retain temporal context information, so they take care of the dependence of the current prediction based on the current input and also the past one. LSTM where proposed by Hochreiter \emph{et. al.} to improve the quality of the long-term memory that these models have by means of gating mechanisms that control the flow of information getting in and out of the network. For further details address \cite{hochreiter1997long}. \\
The RNN sentence embedding method presented by Palangi \emph{et. al.} \cite{palangi2016deep} takes the one-hot representation for each of the words in the text sequences, obtains its word embedding and then feeds the LSTM with them, one at each timestep, keeping the same order as presented in the sequence. The LSTM will update its state based on this embedding and therefore will be accumulating the information of each word and its own context. At the end of this process the LSTM state will have a condensed representation of the whole sentence.

Such dense representations of sequences of text have also been addressed by Cho \emph{et. al.} \cite{cho2014learning} for statistical machine translation with GRU cells, which are a similar approach to that of LSTM. They proposed a RNN architecture called Encoder-Decoder. The first stage encodes a sequence into a fixed-length vector representation and the other decodes the vector into another sequence of arbitrary length. The resulting vector after the encoder stage can be used to represent the sentence.

%%%%%%%%%%%%%%%%%%%%%%%%%%%%%%%%%%%%%%%%%%%%%%%%%%%%%%%%%%%%%%%%%%%%%

\section{Visual Question Answering}
Visual Question Answering is a novel problem for the computer vision and natural language communities, but is has received a lot of attention thanks to the dataset and metrics released with the VQA challenge, together with the large investments of pioneering tech companies such as Microsoft, Facebook or Sony.

The most common approach is to extract visual features of the image using a pretrained off-the-shelf network and process the question using word embeddings or sentence embedding \cite{xiong2016dynamic}\cite{ren2015exploring}\cite{lu2016hierarchical}\cite{kim2016multimodal}\cite{zhou2015simple}\cite{antol2015vqa}\cite{zhu2015visual7w}.

Antol \emph{et. al.} \cite{antol2015vqa}, the organizers of the VQA challenge and the creators of the VQA dataset, propose as their baseline a model that uses VGG-16 \cite{simonyan2014very} to extract the visual features of the image. They use as the representation the output of the last hidden layer of this model. This features are then $l_{2}$ normalized and given to a fully-connected layer to transform the incoming vector into a common space with the question representation. For the question, they use a 2-layer LSTM that takes as the input the word embedding of each question token, timestep by timestep, and when the whole question has been introduced into the LSTM, it outputs its last state as the question embedding. This vector (dimension 2048) is also given to a fully-connected (similarly with what they do with the image) to project it to the same space. Both features are combined using an element-wise multiplication for later use by a fully-connected layer and a softmax that will predict the class answer. Here the 1000 most common answers in the training set have been selected as the classes to predict.

A simple bag-of-words and word embedding model that uses GoogLeNet \cite{szegedy2015going} for the image processing and a concatenation of both visual and textual features is what Zhou \emph{et. al.} \cite{zhou2015simple} present in their paper as a basic approximation to VQA.

\begin{figure}
\centering
\includegraphics[width=1\textwidth]{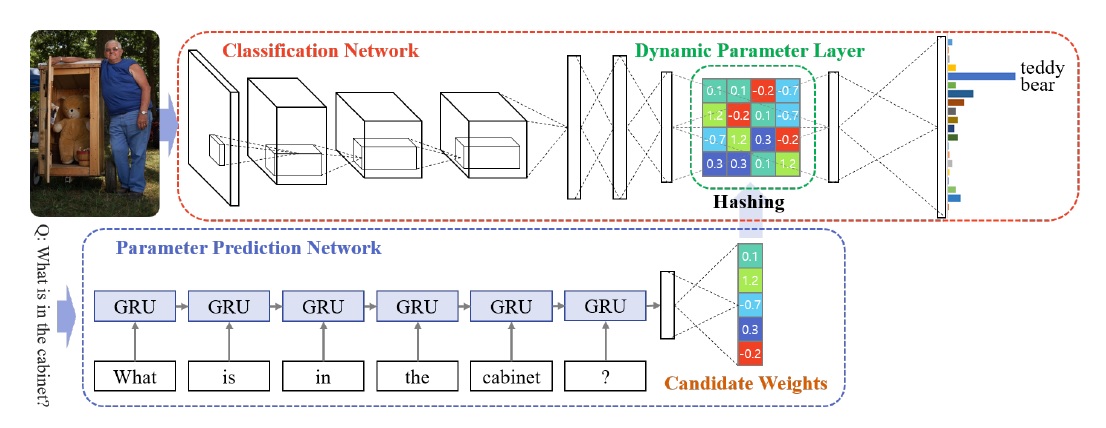}
\caption{DPPnet, the question is used to learn how to predict parameters for a dynamic parameter layer in the classification network.}
\label{fig:dppnet}
\end{figure}

A quite different method from the ones presented above is what Noh \emph{et. al.} \cite{noh2015image} propose, called Dynamic Parameter Prediction Network (DPPnet). They state that in order to solve VQA tasks, different \emph{networks} need to be used as the model need to perform different tasks depending on the question. To accomplish this, they use the question to predict the weights of one of the layers in the networks, thus changing it at test time for each sample. They take the VGG-16 pretrained with ImageNet model as their starting point. Then, they remove the softmax layer and add three fully-connected layers (the last one being a softmax). They have named 'classification network' to this modified VGGnet. The interesting point is that the second one of those new fully-connected layers is a dynamic parameter layer. This means that at test time, the weights of this layer will be changing from sample to sample. These weights are predicted by a network (parameter prediction network) composed by a Gated Recurrent Unit (GRU) \cite{cho2014properties} connected to a fully-connected layer. GRU is another kind of RNN similar to LSTM. This layer takes the embedding of each word in the question as its input and when the whole question has passed through the network, its last state is given to a fully-connected layer which predicts the weight candidates for the dynamic parameter layer in the classification network (the VGG-16 based network). To reduce the number of parameters to predict, a hashing function is used to map from the predicted weights of the prediction network to the actual weights of the fully-connected. Figure \ref{fig:dppnet} depicts the whole architecture of their model.

Other authors propose attention models to improve the performance of the whole model, stating that most of the questions refer to specific image locations \cite{zhu2015visual7w}\cite{xiong2016dynamic}\cite{lu2016hierarchical}. In \cite{zhu2015visual7w} the visual features (the output of fc7 of VGG-16) are treated as if they were the first word in the question, that is fed into a LSTM word embedding by word embedding. The attention model depends on the LSTM state and is used to weight convolutional features of the image (output of the last conv layer of the VGG-16), that are again introduced in the LSTM merged (using addition) with a word embedding. Ren \emph{et. al.} \cite{ren2015exploring} present a similar but simplified method that also treats the image as the first word of the question but that does not have an attention model. Xiong \emph{et. al.} \cite{xiong2016dynamic} present a model based on Dynamic Memory Networks (DNM), that is a modular architecture with attention models. They created a new input module to be able to perform VQA tasks apart from text-based QA and improve the memory module. They use bidirectional GRU so that each feature (textual or visual) has a full context representation thus representing local and global information. 

All these methods present visual attention models but as proposed by Lu \emph{et. al.} \cite{lu2016hierarchical}, attention in the question can also be applied to increase the model performance. This method, called co-attention together with a hierarchical representation of the question helped them to achieve state-of-the-art accuracy with the VQA dataset (and using their evaluation metric). Recently another method has outperformed their results. 

More sophisticated approaches have also been presented, such as Multimodal Residual Learning applied to VQA \cite{kim2016multimodal} that uses Deep Residual Learning to build complex and very deep networks. Other works propose learning methods for specific sub-problems of VQA such as human action prediction and then apply those trained models for VQA tasks \cite{mallya2016learning}.

At writing time, the model that achieves state-of-the-art accuracy is the one proposed by Fukui \emph{et. al.} \cite{fukui2016multimodal} which uses Multimodal Compact Bilinear pooling (MCB) to merge the visual features and the information from the question. They hypothesize that the typical merge actions (addition, element-wise product, concat...) do not express correctly all the information. Using MCB to merge those features they achieve an accuracy of 66,2\% on the Real Open-ended test-dev dataset. A MCB is also used to create two different attention maps that are concatenated before feeding the main MCB.

%%%CHAPTER 3%%%%%%%%%%%%%%%%%%%%%%%%%%%%%%%%%%%%%%%%%%%%%%
\chapter{Methodology}
\label{cha:methodology}

This chapter presents the methodology used to develop this project and the process followed to achieve our final results. 
The baseline for the results that has been taken into account is the one provided by the CVPR16 VQA Challenge.

\section{A programmer's word}
During the first stages of this thesis, when we were looking for some baseline code to perform VQA tasks which we could start with, we found out that the open-sourced projects for VQA were not reusable at all. As a matter of fact, it seems that the vast majority of research code out there has not been developed with programming best practices or with reusability in mind.

That is why we decided to develop our code having in mind some important things: modularity, abstraction, reusability. We intended to apply as many good practices as possible given that we had a restriction in time as we wanted to present our results in the VQA Challenge. As it always happen with software projects, the time variable was crucial in terms of \emph{how much} modular, abstract or reusable was our code at the end. 

Nevertheless, we think that the final work is going to be very useful as a starting point for future projects related with VQA and also as a good end-to-end Keras' example. With that we mean that sometimes there is a lack of examples with some degree of complexity that cover the whole process of building a model, training, validating and testing.

\subsection{Choosing the best language}
In terms of actually coding the project we decided to use Python as a programming language. We considered this was the best language to approach this project in terms of prototyping speed and tools available. \\
C++, Lua and Python were the finalists of this search. C++ was discarded as sometimes is a little bit cumbersome to prototype things fast, for the syntax itself and for the fact that it is a compiled language. Lua and Python have a quite similar syntax, both being a high-level and scripting programming language with a fast learning curve and fast for prototyping. At the beginning, all the open-sourced projects that we found that had something to do with VQA where written in Lua using a deep learning framework called Torch\footnote{\url{http://torch.ch/}}. This seemed a good reason to choose Lua over Python but then, having a look into where the community was going to, we found out that frameworks like Theano or TensorFlow were having great success and the developers and research community was moving towards them. Both frameworks are for Python, which made us choose Python as a programming language for the project. Then, we were recommended to use Keras, a library able to work upon Theano or TensorFlow to expedite the prototyping process.

\subsection{A Pythonic project}
Having chosen Python as the programming language, one of the things we wanted to do to increase the readability and reusability of this project was to follow a code style guide. 

In the programming world there are many languages and for everyone of them there are tons and tons of styles that the developers tend to program with, and we programmers are picky. That is why the most popular languages usually have a code style guide that define \emph{how the code should look} and \emph{what is a good practice} in that language. Using these code style guidelines increases the readability of the code and helps you to develop a better code that will be more easily extended or used.

For Python, this code style guideline is called PEP8\footnote{\url{https://www.python.org/dev/peps/pep-0008/}}\footnote{There are different PEP guidelines, each one of them dictating best practices for different tasks}. The code presented with this project follows the PEP8 guideline.

\subsection{An eight-leg cat called octocat}
To develop the project as professional as possible and to keep track of the changes we made we have used Git\footnote{\url{https://git-scm.com/}} as a Version Control System (VCS). Git give us the ability to work in parallel when needed and to prototype things without the fear of not being able to restore our previous work or having to do all those annoying manual backups. Using Git we have created an historic of our project development process.

To store our Git repository we have used GitHub as it allows us to open-source the project once finished and to enable community contributions. After the VQA challenge deadline we published our GitHub repository\footnote{\url{https://github.com/imatge-upc/vqa-2016-cvprw}} as public so everyone can use the code.

%%%%%%%%%%%%%%%%%%%%%%%%%%%%%%%%%%%%%%%%%%%%%%%%%%%%%%%%%%%%%

\section{Dataset}
\label{sec:dataset}

In order to train a model (in supervised learning) we need a very large amount of data. This data are example of input-output pairs. In our case, the input are both the image and the question and the output is the answer. 

To train our models we have used the real image VQA dataset\footnote{\url{http://www.visualqa.org/download.html}}, which is one of the largest visual question-answering datasets. This dataset is provided by the organizers of the VQA Challenge and is splitted in the typical three subsets: train, validation and test. The train subset is composed by 82.783 images, 248.349 questions and 2.483.490 answers; the validation by 40.504 images, 121.512 questions and 1.215.120 answers; and finally the test set is composed of 81.434 images and 244.302 questions. The whole explanation on how the organizers created this dataset can be found in their paper \cite{antol2015vqa}.

All the images are part of the Microsoft Common Objects in Context (MS COCO) image dataset\footnote{\url{http://mscoco.org/}} and the questions and answers have been generated by different workers. MS COCO was chosen as the image are very different (size, color and black\&white, quality...) and they are rich in contextual information. The questions are of different types, being the most common the 'what...?', 'is...?', 'how...?' and 'are...?'.

As it is usual, the train subset has been used to learn the model parameters, at the same time that the validation set was used to check on the model's generalization to unseen data. By using this information, we could improve some parameters and present the ones that achieved higher accuracy on the validation set. The test set does not have answers as it defines the problem to be solved during the challenge. The answers predicted for the test set were the ones that we submitted to the VQA 2016 challenge.

\section{Text-based QA toy example}
As we already stated in our work plan \ref{sec:workplan}, we started familiarizing with VQA tasks and how Keras library works through a text-based QA model. 

The kind of text-based QA problem that we addressed was a toy example were a short story and a question related to that story is given to the model so it can predict a single word answer.

\subsection{Tokenization}

The first step to take is transforming the words (from the story and the question) into numbers that can be feed into the model. We did this preprocessing with a tokenizer provided by Keras that is in charge of tokenizing the text sequences. By tokenizing here we mean splitting the whole string into words, remove the unnecessary ones (punctuation for example) and transform each word into a number. This number is the index of that word in a dictionary that we created previously. The dictionary or vocabulary of our tokenizer can be a predefined one or not. We did not use a predefined dictionary but created our own one using the training data. \\
To create such a dictionary, its size is important, the number of unique words that it can include. A special word will also be included, which is the one representing 'unknown' words, \emph{i.e.} words that are not in the previous list. \\
From this point on, a word is no longer a string such as 'garden' but a number representing its position in the dictionary.

\subsection{Model architecture}
\begin{figure}
\centering
\includegraphics[width=1\textwidth]{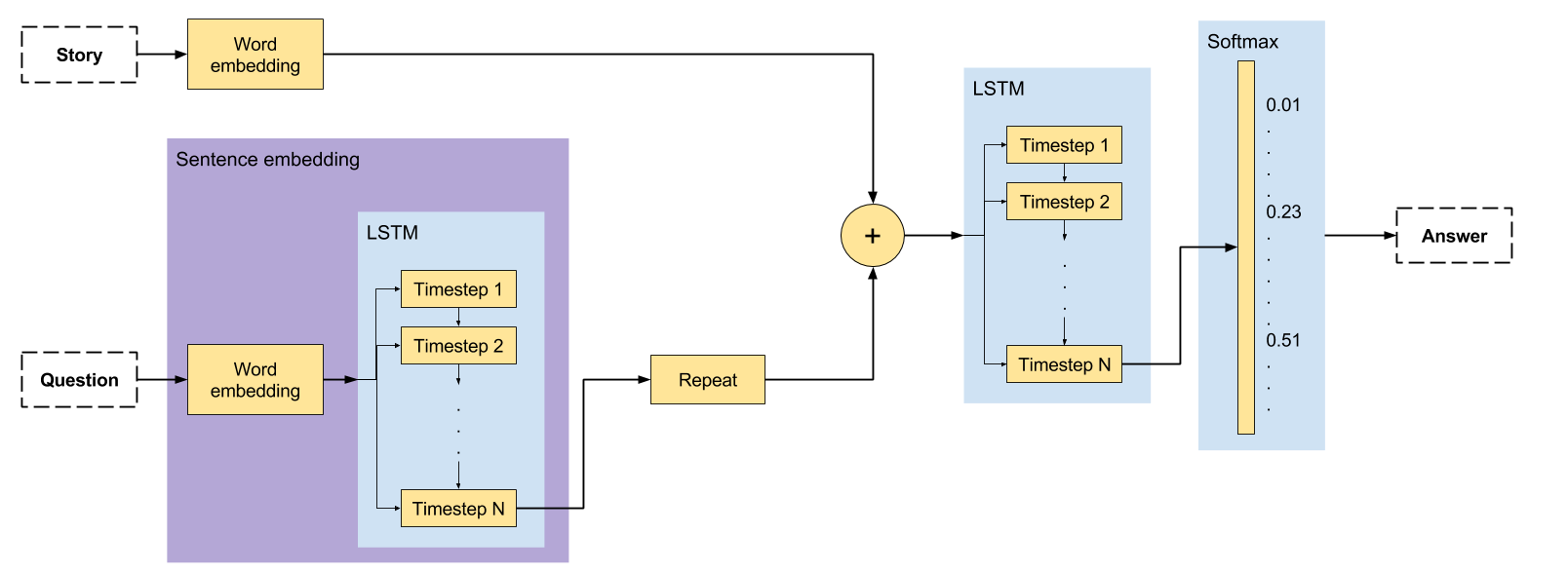}
\caption{Schema of the model used for text-based QA}
\label{fig:text-based-qa-schema}
\end{figure}

Once we have tokenized the story and the question, their representation is a list of numbers. These numbers are the input of our model. The architecture of this model is represented in Figure \ref{fig:text-based-qa-schema}. The model has two branches, one for the story and the other one for the question, that are merged together to produce the output. 

This model has an encoder structure (see Cho \emph{et. al.} \cite{cho2014learning} for a complete encoder-decoder architecture for automatic machine translation). We take the input, which is high dimensional (the size of the vocabulary, a typical value is 20.000) and we encode it into a much smaller representation, a vector in a continuous space of a dimension that we have to specify beforehand (it is fixed and it is an hyper-parameter that we have to adjust manually), for example 100. This single vector holding the whole information of the question and the story is our encoding. This encoded vector is given to the fully-connected layer, a softmax, that will predict the one-hot representation of the answer. A one-hot representation is simply a vector with all zeros and just a one in a specific location. In our case, this vector has dimension equal to the vocabulary size and the one is placed in the location equal to the word's index in the dictionary. The softmax will not predict 0 and 1 but a distribution between 0-1. We took the maximum value as our output. 

Lets dig a little bit deeper into the encoder. The story branch has only a word embedding block \cite{mikolov2013distributed}. The word embedding learns how to do a dense representation of the input word as a vector in a continuous space of the specified dimension. This reduces the dimensionality of our input as it is projected into a space with less dimensions. Such space has a very interesting property which is one of the reasons that we use these embeddings. The vectors in that space are not only dense but they are also a semantic representation of the word. One possible example of this is that the embeddings of words with similar meaning are close to each other (the distance between the vectors is small).
After the word embedding we will have a sequence of vectors representing the story, a vector for each word.

The question branch is a little bit more complex. We start with the same word embedding than the story. The output of that block is then given to a LSTM \cite{hochreiter1997long}, which is a Recurrent Neural Network (RNN). RNNs have the advantage of having memory over \emph{time}, \emph{i.e.} they have a state that is kept in memory and it is updated in each iteration and their output is somehow dependent of this state. LSTMs are widely used to process sequences for these reasons. Specifically, we are using a non-stateful LSTM that means that this state is not preserved from batch to batch, it is resetted. We have also configured the LSTM so it only outputs its last state. We set the maximum sequence length to the LSTM so it knows when the question has finished and it can output a value.

The output of the LSTM is a representation in a single vector of the whole question. This vector is then given to a block that repeats the vector as many times as specified, in our case, the maximum story length (in tokens). This combination of the word embedding and a LSTM that sees all the question words and then outputs its memory state is known as a sentence embedding \cite{palangi2016deep}.

This way, at the merge stage there is a sequence of word embeddings from the story branch, and a sequence of the question embedding repeated. Each iteration in the sequence is what we call a \emph{timestep}. That being said, at each timestep we are summing up the embedding of a story word and the embedding of the whole question. To be able to do so, both vectors need to have the same dimension and that forces that both word embeddings (story and question) must have the same dimension as hidden units in the LSTM that encodes the question (which determines the output dimension of it).

The result of adding both embeddings is given to another LSTM which is in charge of the last encoding. This LSTM is also non-stateful and it will accumulate all the merged features until it has seen the whole story and then it will output its state. This last vector is our encoding of the whole story and question merged and it is what we use to predict the answer, as we have explained before.

As an addition, this model also uses drop outs with a 0,3 value to prevent overfitting.

\subsection{Model up and running}
Once we had this model built, we trained it with 10.000 sample for the QA1 task defined by Weston \emph{et. al} \cite{weston2015towards} and we did some small tests. We did not run extensive experimentation at this stage as this was not our objective and because the parameters of the model and the dataset we used were designed more as a toy example than a real-world solution. This stage allowed us to train the model as fast as possible and check that the whole process was working.

%%%%%%%%%%%%%%%%%%%%%%%%%%%%%%%%%%%%%%%%%%%%%%%%%%%%%%%%%%%%%%%

\section{Extending the text-based QA model for visual QA}
\label{sec:first-vqa}

Taking as a starting point the previous text-based QA model, we modified it so it could be used for visual QA. Notice that the architecture shown in Figure \ref{fig:text-based-qa-schema} has been built around the idea that we have a story that gives some information to the model and then we ask a question about that story. The model uses the information retrieved from the story to be able to answer the question. In visual QA our story is the image, is what give us the information needed to answer the question.

\subsection{Image processing}

With the idea of our image being the "story" from which we have to extract information, we changed the story branch for an image branch. In such a branch we use the VGG-16 convolutional neural network proposed by Simonyan \emph{et. al.} \cite{simonyan2014very}, an off-the-shelf model, to extract the visual features, as you can see in Figure \ref{fig:visual-qa-1-schema}. We did not use the output of the whole model but we truncated it until the last convolutional layer, before the fully-connected fc-4096. Using the output of the conv layers instead of the fully-connected ones is a common practice to extract visual features maps. 

In order to be able to combine this visual information with the one obtained from the question, we need to turn this 2D map into a vector. We used a Flatten layer to do so and then we give this vector to the repeat block. Notice that now we are repeating the image (our visual story) instead of the question. We are doing this as the question is our only sequence in this model and this way the model will \emph{see} the whole image for each question word.

\subsection{Modifications on the question branch}
\begin{figure}
\centering
\includegraphics[width=1\textwidth]{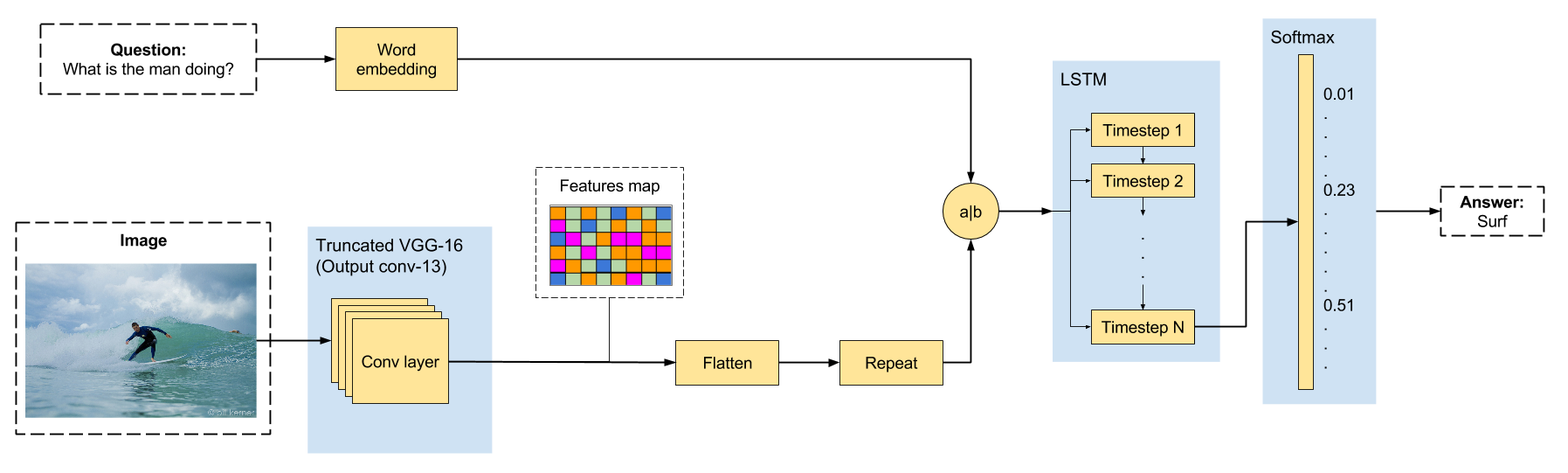}
\caption{First visual QA model's schema}
\label{fig:visual-qa-1-schema}
\end{figure}

As shown in Figure \ref{fig:visual-qa-1-schema}, the question branch has only a word embedding now. This means that in each timestep a question word will be feed into the model and, because the visual features are repeated, each one will be merged with the information of the whole image. 
The dimension of the word embedding and the visual features is different so our merge process now is not a summation but a concatenation of both vectors.

\subsection{Model parameters}
The important parameters of this model are: vocabulary size, LSTM hidden units, embedding size, question maximum length, learning rate and batch size. It is also important which optimizer to use.

We set the batch size to be the maximum that we could save in the GPU RAM, having a value of just 32 samples. We need to consider that we also need to fit the compiled model (its weights) in the GPU RAM and this is very expensive as some of our layers, and thus its weights, are huge as we will see now. 

The learning rate for the network's parameter was governed by the Adam optimizer \cite{kingma2014adam} which modifies the learning rate through the training. We only need to specify the starting learning rate, and we chose the default one proposed by Kingma \emph{et. al.} in the original paper which is 0,001.

For the question maximum length we have taken the length of the largest question in the training subset. This parameter is used in the last LSTM so it knows when it has seen the whole question and can output its state. We found that for the training set, the maximum question length is 22. The questions that have a smaller length have been left-padded with 0, so the input is 'inactive' and then it is activated with the question tokens. The network has been configured to ignore these padding zeros.

The vocabulary size is crucial for the softmax layer as this will set the number of neurons of this layer. A value of 20.000 was chosen as it is quite common and respects the tradeoff between number of words (which give more flexibility to the model) and number of weights to train (time consuming, training problems, memory constraints).

For this model we chose the number of LSTM hidden units and the embedding size to be the same, with a value of 100. We used this value for simplicity and due to some experience of a team member regarding these parameters.

We have also changed the dropout rate from 0,3 to 0,5.

\subsection{Implementation and training}
\label{sec:implementation-training}
We built the whole model including the truncated VGG-16 and we used pretrained weights for this module, that we froze at training time. This weights were the result of training the whole VGG-16 on ImageNet\footnote{\url{http://www.image-net.org/}} one of the biggest image datasets existing nowadays. As we freeze the VGG-16 weights we did not make a fine-tunning of it, we only trained our own layers.

To train this model we started making use of the computational service of the Image Processing Group (GPI) at the Universitat Politecnica de Catalunya. Some memory constraints were faced as the size of the vocabulary imposed the need to create huge vectors representing the answers and the size of the compiled model reflected this too.
As we could only fit 32 samples per batch, the training process was at a rate of 17-24 hours per epoch using NVidia Titan X GPUs, equipped with 12 GB of RAM. This forced an evolution to the next model as having this model train for a reasonable number of epochs (40) was not a valid option. Notice than an epoch is defined as a single pass of all the examples in the training set through the model under training.

In terms of software, we created an Image, Question, Answer, VQASample and VQADataset Python classes to hold the information of these entities and to allow single responsability and modularity of the code. Using these classes we encapsulate the logic in modules that we can easily control, instead of working with plain data such as Python dictionaries, lists, numpy arrays, etc.

%%%%%%%%%%%%%%%%%%%%%%%%%%%%%%%%%%%%%%%%%%%%%%%%%%%%%%%%%%%%%%%%%%%%

\section{Model improvement: towards the final model}
\label{sec:kcnn-model}

The prohibitive duration of the training process made us opt for precomputing the visual features of the image. This approach made sense as we were not modifying the values of the VGG-16 convolutional network that was in charge of extracting these features.

Instead of precomputing ourselves the visual features using an isolated VGG-16, our partners from the Computer Vision group at the Universitat de Barcelona (UB) provided us with these features extracted with a new kind of CNN called Kernelized CNN (Liu \emph{et. al.} \cite{liu2015kernelized}). You can find a short description in the Methods and procedures section \ref{sec:methods-procedures}. The dimension of the output vector from the KCNN module is 1024. The rest of the parameters and functionality remains the same as in the architecture described in Section \ref{sec:first-vqa}.

\begin{figure}
\centering
\includegraphics[width=1\textwidth]{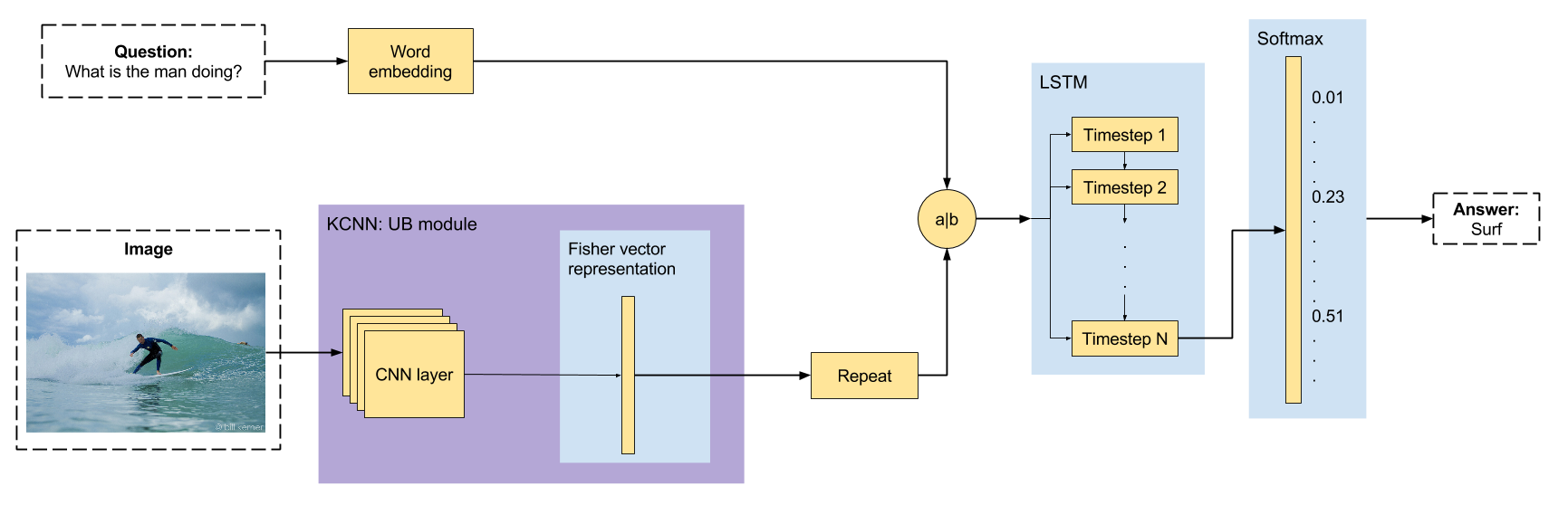}
\caption{Replacement of the VGG-16 by the KCNN model}
\label{fig:visual-qa-2-schema}
\end{figure}

\subsection{Implementation and training}
We also changed the way we programmed the model. In the previous case, we were using an abstract model implementation of Keras called Sequential\footnote{\url{http://keras.io/getting-started/sequential-model-guide/}} which is basically a stack of layers. This model also allows the possibility of merging two sequential models into one, that is what we used to create the two input branches. For this modified model we changed to the more flexible Functional API\footnote{\url{http://keras.io/getting-started/functional-api-guide/}} which is thought to build more powerful models in a graph approximation. This new interface let us work with the tensors themselves so it is now easier to modify and make the model more complex.

By using the precomputed visual features and this new implementation, we reduced the training time of an epoch to less than an hour (40 minutes approximately).

\subsection{Batch normalization and reducing the learning rate}
\begin{figure}
\centering
\includegraphics[width=1\textwidth]{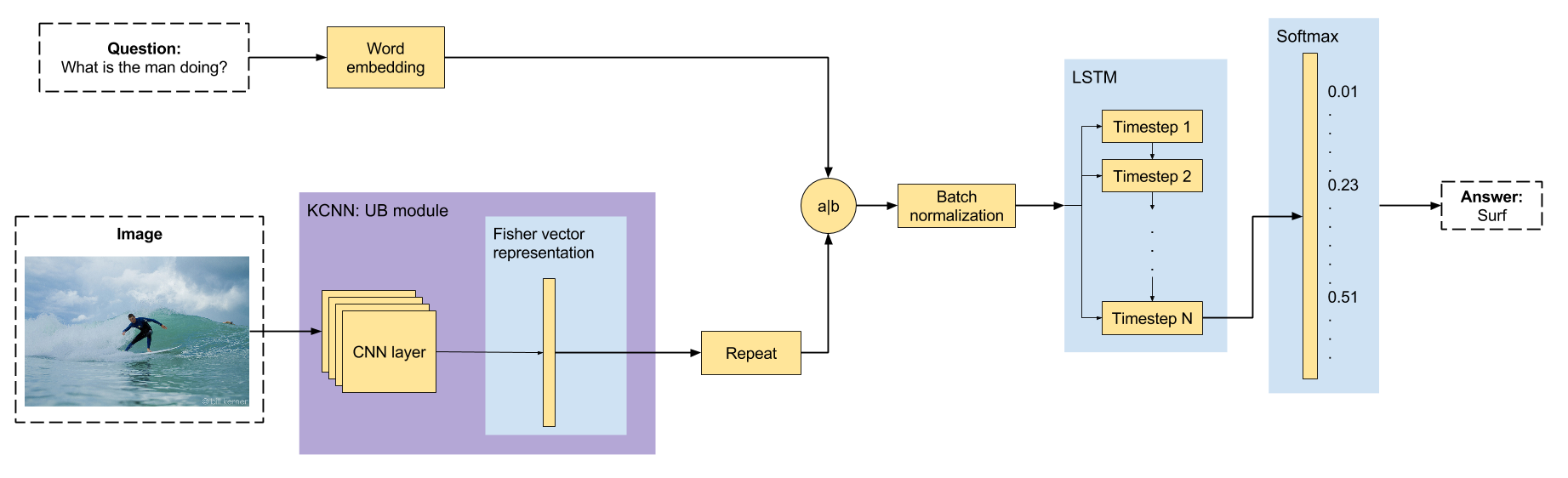}
\caption{Batch normalization added to help the training process}
\label{fig:visual-qa-3-schema}
\end{figure}

One of the first modifications we tried was adding a batch normalization  layer (Figure \ref{fig:visual-qa-3-schema}) after the merging process, as this helps the training process and usually increases the accuracy. Ioffe and Szegedy propose to introduce the normalization of the layers' input distribution inside the model architecture \cite{ioffe2015batch}. They introduce this normalization using their novel layer (batch normalization) which reduces the internal covariate shift.

We also reduced the initial learning rate sequentially from 0,001 to 0,0003 and to 0,0001 and we found that the last one was giving the best accuracy as we will explain later in the results chapter \ref{cha:results}.

%%%%%%%%%%%%%%%%%%%%%%%%%%%%%%%%%%%%%%%%%%%%%%%%%%%%%

\section{The final model: sentence embedding}
\label{sec:final-model}

\begin{figure}
\centering
\includegraphics[width=1\textwidth]{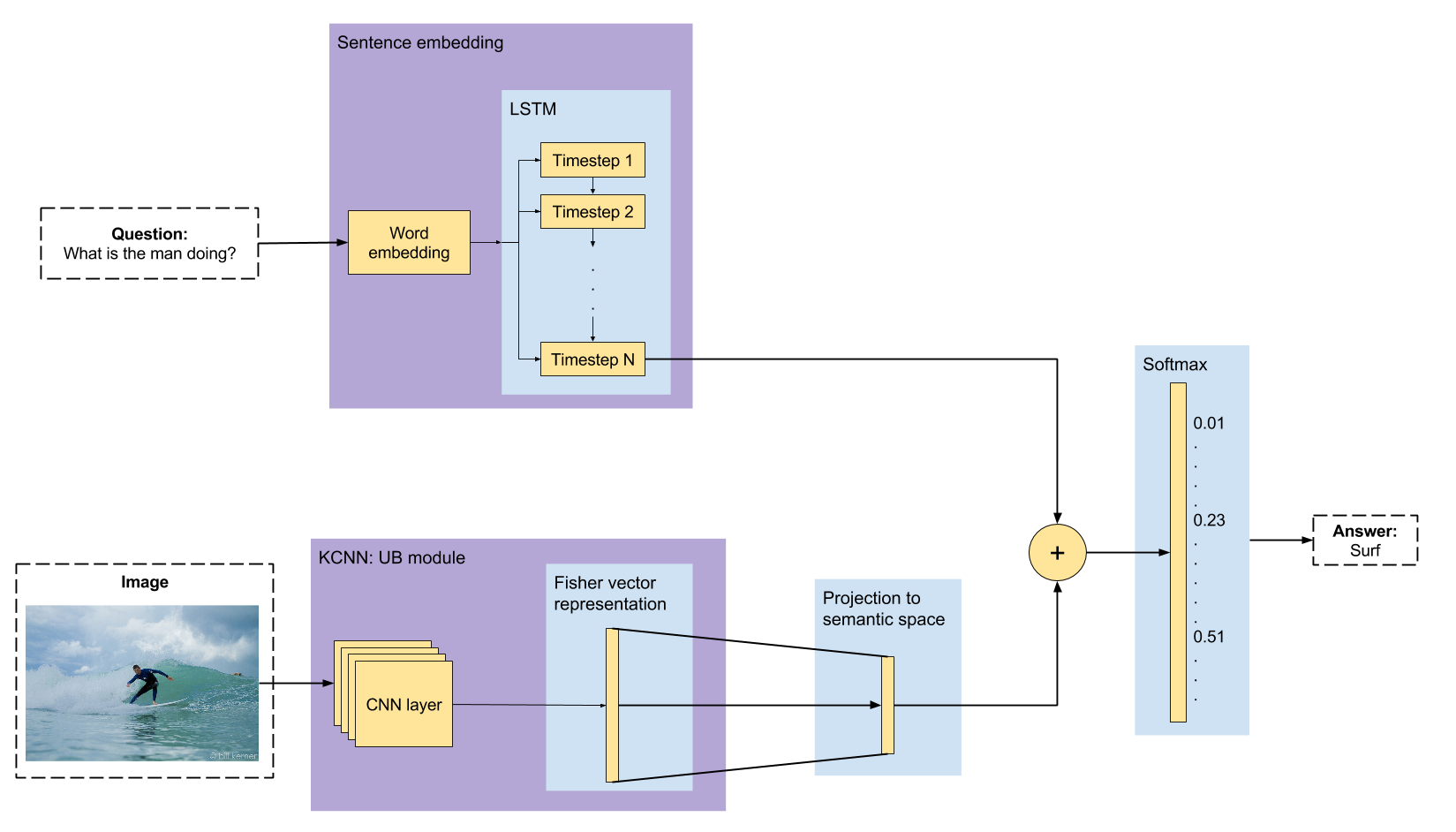}
\caption{Final model. A sentence embedding is used for the question and the visual features are projected into the same semantic space than the question}
\label{fig:visual-qa-4-schema}
\end{figure}

Our last model was the one that predicted the answers with higher accuracy and presented to the VQA challenge. Several changes were introduced with respect to the preliminary prototypes so lets have a look into the different blocks, depicted in Figure \ref{fig:visual-qa-4-schema}.

\subsection{Question embedding}
The question branch was modified by adding a LSTM at the end of the word embedding, thus creating a sentence embedding, in our case the question embedding. The resulting vector of the sentence embedding module is a dense and semantic representation of the whole question as it was in our text-based QA model \ref{fig:text-based-qa-schema}. The difference here is that we did not choose the same value for the word embedding dimension and the number of LSTM hidden units. We set 100 as the word embedding dimension and 256 as the number of LSTM hidden units, which is a common value. We increased the number of hidden units as this can help increasing the accuracy in the condensed representation of the questions but we did not change the embedding dimension as this could decrease the density of the word embedding representation.

\subsection{Image semantic projection}
We decided to add a fully-connected layer after the KCNN module to be able to project the visual features into a space of the same dimension as the question embedding. The fully-connected layer can be seen as a matrix operation which projects the features' 1024-vector into a 256-vector in the semantic space. We have chose ReLU as the activation function for this layer.

\subsection{Merging and predicting}
As both textual and visual features were projected into a 256-dimensional space, we can sum up them together and merge these features.
Now that both the question and image are represented by a single vector and not by a sequence of vectors, there is no need to add a LSTM after the merge and we can feed the resulting merged vector to the softmax so it can predict the answer.

\subsection{Other modifications}
The learning rate of this model was initialized to 0,0001 against the 0,001 of the first KCNN model.
We also tried to reduce the learning rate to 0,00001 and to add a batch normalization stage after the merging process but as we will see in the following chapter, neither of those increased the accuracy of the original final model. \\
Before submitting to the VQA challenge over the test set, we also tried to train the model with the whole training subset and the 70\% of the validation subset but this did not help either.

%%%CHAPTER 4%%%%%%%%%%%%%%%%%%%%%%%%%%%%%%%%%%%%%%%%%%%%%%
\chapter{Results}
\label{cha:results}

In this chapter, the results of the different models exposed in the Methodology chapter \ref{cha:methodology} will be presented. 

\section{Evaluation metric}
The models have been evaluated using the metric introduced by the VQA challenge organizers. As they state in their evaluation page\footnote{\url{http://www.visualqa.org/evaluation.html}}, this new metric is robust to inter-human variability in phrasing the answers.

The new accuracy formula per answer is the following:
\begin{equation}
Acc(ans) = min\left(\frac{\mbox{\#humans that said ans}}{3}, 1\right)
\label{eq:accuracy}
\end{equation}

The accuracy over the whole dataset is an average of the accuracy per answer for all the samples.

The interpretation of equation \ref{eq:accuracy} is as follows: an answer is given as correct (accuracy equals 1) if the same exact answer was given by at least three human annotators. Zero matches equals zero accuracy and from there each match gives 0,33 points to the accuracy with a maximum of 1.

%%%%%%%%%%%%%%%%%%%%%%%%%%%%%%%%%%%%%%%%%%%%%%%%%%%%%%%%%%%%

\section{Dataset}

At this point is worth summarizing the dataset characteristics mentioned in \ref{sec:dataset}:
\begin{itemize}
\item \textbf{Training dataset}: 82.783 images, 248.349 questions, 2.483.490 answers
\item \textbf{Validation dataset}: 40.504 images, 121.512 questions, 1.215.120 answers
\item \textbf{Test dataset}: 81.434 images, 244.302 questions
\end{itemize}

Notice that for each image there are three questions and for each question there are ten answers. These ten answers were provided by human annotators and the most frequent ones were selected. Most of the answers are the same but rephrased.

The organizers also provide a Python script to evaluate the results\footnote{\url{https://github.com/VT-vision-lab/VQA/}} the same way that they do when you submit the test results. This script preprocess the answers to avoid format-like mismatches. As an example, they make all characters lowercase, remove articles, convert number words to digits...

This script needs a specific JSON file with the ground truth answers and another one with the machine generated answers (what the model has predicted) in a predefined format.
We used this script to perform an evaluation of our model over the validation set (because we do have the answers for this subset).

%%%%%%%%%%%%%%%%%%%%%%%%%%%%%%%%%%%%%%%%%%%%%%%%%%%%%%%

\section{Models architectures and setups}

In the following section we will refer to the models by a number in order to be more clear and concise. These identifiers are defined here with a description of the model/configuration:

\begin{table}[h]
\centering
\begin{tabular}{| p{1.5cm} | p{11cm} |} 
\hline
	\textbf{Identifier} & \textbf{Description} \\
\hline
	0 & First VQA model. Uses VGG-16 to extract visual features. Described in \ref{sec:first-vqa} and schema in Figure \ref{fig:visual-qa-1-schema} \\
\hline
	1 & Improvement of model 0 using KCNN. Described in \ref{sec:kcnn-model} and schema in Figure \ref{fig:visual-qa-2-schema} \\
\hline
	2 & Model 1 using with batch normalization. Schema in Figure \ref{fig:visual-qa-3-schema} \\
\hline
	3 & Model 1 with a learning rate of 1/10 the original (0,0001) \\
\hline
	4 & Final model using sentence embedding for the question and a projection to the semantic space for the image. Described in section \ref{sec:final-model} and schema in Figure \ref{fig:visual-qa-4-schema} \\
\hline
    5 & Model 4 using batch normalization \\
\hline
\end{tabular}
\caption{Models identifiers and descriptions}
\label{tab:models}
\end{table}

Results for model 0 will not be presented as we only completed the building stage but we did not finish the training process for the problems already explained in Chapter \ref{cha:methodology}. We only include it here to state that this was our base VQA model.

%%%%%%%%%%%%%%%%%%%%%%%%%%%%%%%%%%%%%%%%%%%%%%%%%%%%%%%%%%%%%

\section{Training and validation losses}

One of the earlier results that helped us to improve our models was the training, and most important, the validation loss. In the following figures you can see the evolution of the training and validation loss per epoch

\begin{figure}[H]
\centering
\begin{minipage}{.5\textwidth}
  \centering
  \includegraphics[width=1\textwidth]{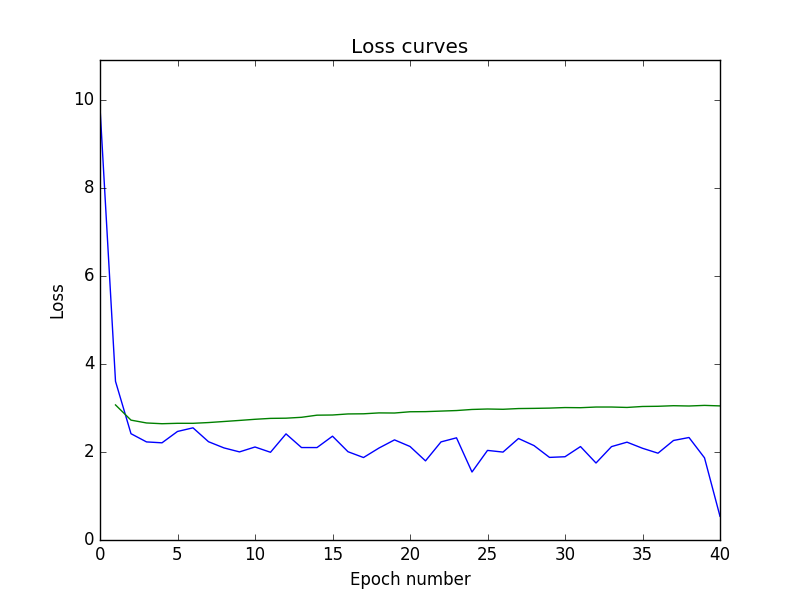}
\caption{Training losses (blue) and validation losses (green) for model 1}
\label{fig:loss-model-1}
\end{minipage}%
\begin{minipage}{.5\textwidth}
  \centering
  \includegraphics[width=1\textwidth]{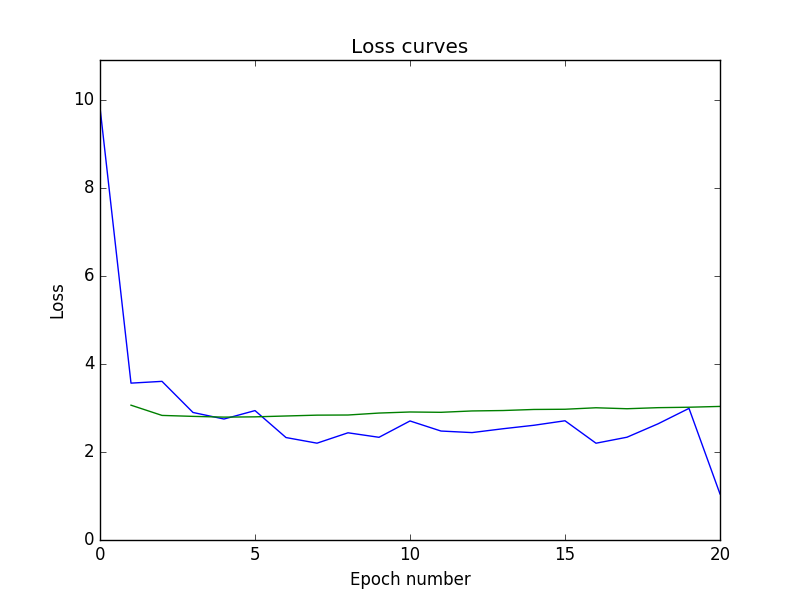}
\caption{Training losses (blue) and validation losses (green) for model 2}
\label{fig:loss-model-2}
\end{minipage}
\end{figure}

For model 1 and 2 we have that the validation loss increases from epoch 3 until the end. We can also appreciate that after the second epoch the model is not learning anymore, the training loss gets stucked around a fixed value with some "noise" and also that in the first epoch the model experiments a huge decrement in the training loss. Both factors are an indicator that the models are slowly diverging and thus the learning rate is too high.

\begin{figure}[H]
\centering
\includegraphics[width=0.8\textwidth]{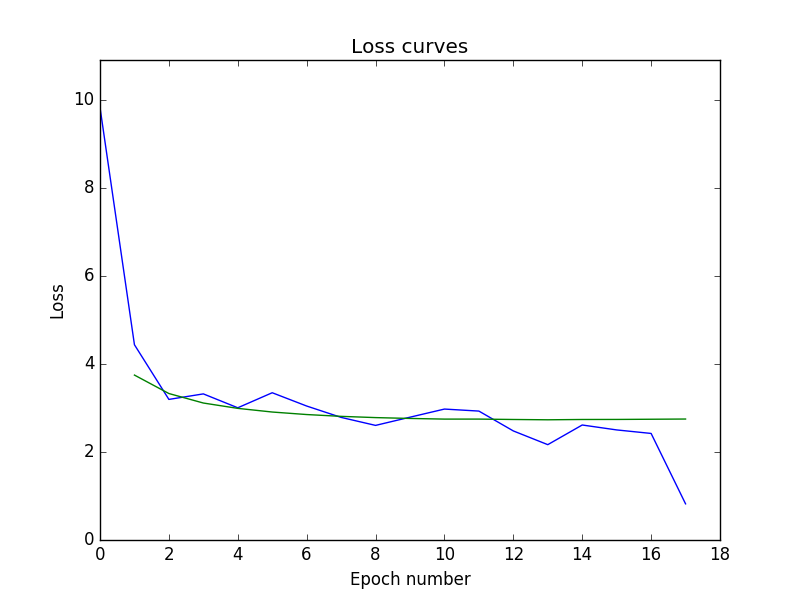}
\caption{Training losses (blue) and validation losses (green) for model 3}
\label{fig:loss-model-3}
\end{figure}

In the model 3 we decrease the learning rate to 1/10 of the original one, having a value of 0,0001. As we can easily see in the plot in Figure \ref{fig:loss-model-3} we experiment a slower decrease in the training loss and it does not stop learning after the first epochs. Even if the average training loss is higher than in the previous models, the validation loss (which is the one that helps us measure how good our model generalizes to unseen data), is lower in this model and does not increase over the iterations.

\begin{figure}[H]
\centering
\begin{minipage}{.5\textwidth}
  \centering
  \includegraphics[width=1\textwidth]{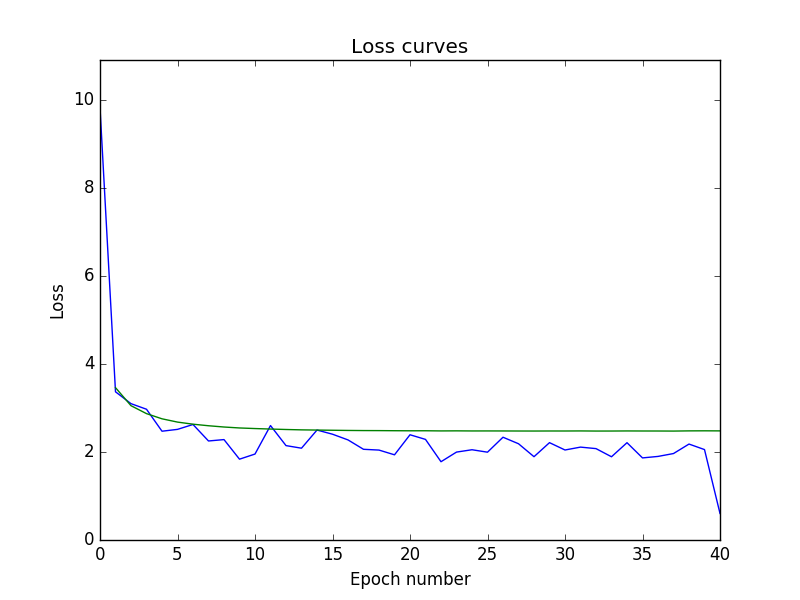}
\caption{Training losses (blue) and validation losses (green) for model 4}
\label{fig:loss-model-5}
\end{minipage}%
\begin{minipage}{.5\textwidth}
  \centering
  \includegraphics[width=1\textwidth]{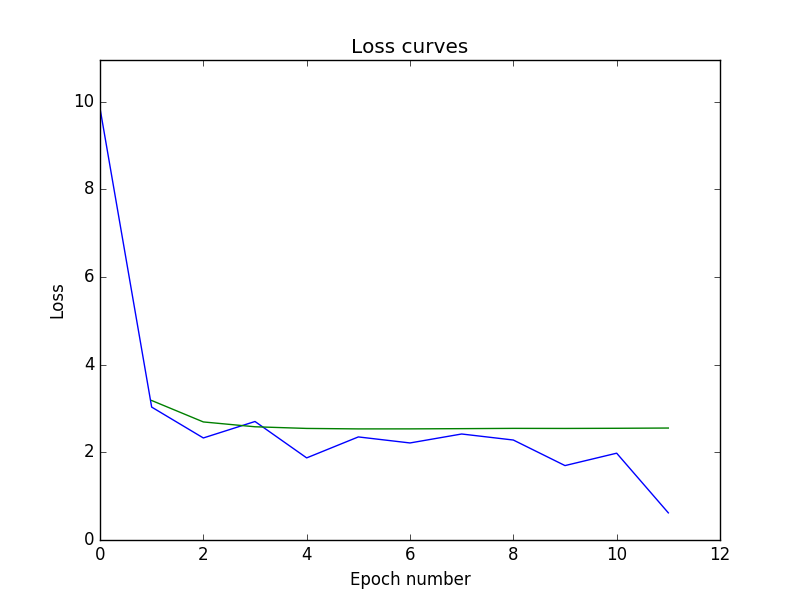}
\caption{Training losses (blue) and validation losses (green) for model 5}
\label{fig:loss-model-6}
\end{minipage}
\end{figure}

Changing to a sentence embedding and projecting the visual features to the same space than the question reduced the validation loss. Having a look at Figure \ref{fig:loss-model-5} we can see that the validation loss slowly decreases epoch after epoch and it reaches the lowest value of the past models. 

Adding a batch normalization layer did not help us in order to obtain better results with the model 5 (Figure \ref{fig:loss-model-6}).

%%%%%%%%%%%%%%%%%%%%%%%%%%%%%%%%%%%%%%%%%%%%%%%%%%%%%%%%%%%

\section{Quantitative results in the VQA Challenge 2016}

The model we presented to the CVPR16 VQA Challenge was the model number 4. We get an accuracy of \textbf{53,62\%} over the test dataset. In table \ref{tab:results} we present a comparison between our accuracy and the accuracy of the baseline model and the top one.

As we did not submit all the results from the different models, we do not have test accuracies for some of them (model 2 and 3).

\begin{table}
\makebox[\textwidth][c]{
\begin{tabular}{| p{3cm} | r | r | r | r | r | r | r | r |} 
\hline
	& \multicolumn{4}{|c|}{\textbf{Validation set}} & \multicolumn{4}{|c|}{\textbf{Test set}} \\
\hline
	\textbf{Model} & \textbf{Yes/No} & \textbf{Number} & \textbf{Other} & \textbf{Overall} & \textbf{Yes/No} & \textbf{Number} & \textbf{Other} & \textbf{Overall} \\
\hline
	Baseline All yes & - & - & - & - & 70,97 & 0,36 & 1,21 & 29,88 \\
\hline
	Baseline Prior per question type & - & - & - & - & 71,40 & 34,90 & 8,96 & 37,47 \\
\hline
	Baseline Nearest neighbor & - & - & - & - & 71,89 & 24,23 & 22,10 & 42,85 \\
\hline
	Baseline LSTM\&CNN & - & - & - & - & 79,01 & 35,55 & 36,80 & 54,06 \\
\hline
	UC Berkeley \& Sony & - & - & - & - & 83,24 & 39,47 & 58,00 & 66,47 \\
\hline
    Humans & - & - & - & - & 95,77 & 83,39 & 72,67 & 83,30 \\
\hline
\hline
	Model 1 & 71,82 & 23,79 & 27,99 & 43,87 & 71,62 & 28,76 & 29,32 & 46,70 \\
\hline
	Model 3 & 75,02 & 28,60 & 29,30 & 46,32 & - & - & - & - \\
\hline
	Model 2 & 75,62 & 31,81 & 28,11 & 46,36 & - & - & - & - \\
\hline
	Model 5 & 78,15 & 32,79 & 33,91 & 50,32 & 78,15 & 36,20 & 35,26 & 53,03 \\
\hline
	Model 4 & \textbf{78,73} & \textbf{32,82} & \textbf{35,5} & \textbf{51,34} & \textbf{78,02} & \textbf{35,68} & \textbf{36,54} & \textbf{53,62} \\
\hline
\end{tabular}
}
\caption{Results of the four baselines provided by the VQA challenge organizer, the state-of-the-art (UC Berkeley \& Sony) and our five models. Model 4 was the one submitted to the challenge leaderboard as the results of our team}
\label{tab:results}
\end{table}

\subsection{A general overview}
The first interpretation of these results is that the gap between the accuracy of the baseline model and the best one (UC Berkeley \& Sony) is quite small, only a 12,41\%. What this means is that it is very hard to create models good at solving visual QA as the model needs to have a deep understanding of the scene and the question and also quite good reasoning abilities. 

Another fact to notice is that there is a performance difference between humans and models performing such tasks, and that means that there is still space for further research in this area.
Related with this, it is worth mentioning that human accuracy using this metric is quite low, comparing with what one would expect it to be (close to 1). This may imply that the metric used to evaluate this tasks may not be the best one to use as it does not reflect correctly the performance in such tasks. This could also be a problem on how the dataset is built. If we check the human accuracy using the metric and dataset presented by Zhu \emph{et. al.} \cite{zhu2015visual7w} we can see that it is 96\%, much more logical a priori.

\subsection{Our results}
Now, if we evaluate our results we can see in table \ref{tab:results} that our model performs slightly worse than the baseline provided by the VQA challenge organizers. This has a reason underneath it. 

The first reason is that our model predicts only single word answers. This means that we will not have a 100\% accuracy in multi word answers as we will never have a complete match. It is true that the VQA dataset's answers are mostly single word but it is also true that we already start with fewer accuracy due to this fact. The VQA dataset answer average length is 1,1 word with a deviation of 0,4.

The second and most important reason is that the baseline model and many of other models presented in the challenge (\cite{lu2016hierarchical}, \cite{noh2015image}, \cite{ma2015learning}, \cite{fukui2016multimodal}, \cite{kim2016multimodal}, \cite{zhou2015simple}, \cite{antol2015vqa}), including the top ones, are a classifier built upon the N (in their case 1000) most frequent answers. This means that they were taking the 1000 answers from the training set that appear more frequently and assign a class label to each one of them. Then, they train their model to learn how to predict which one of these classes is the answer. At test time they predict a class label that it is matched with a predefined answer. Notice that their classes are the whole answer, not words. What this implies is that the model can only predict some of the answers that it has already seen in the training subset but it can not generate new ones, thus being very limited to this specific dataset.

In contrast, our model was built with the idea of being able to generate any word of the vocabulary as an answer, even if during training time that word was never used as an answer. We accomplished that by having the model output a word instead of a predefined answer. As our model has an encoder structure, we could also attach at the end a decoder stage to predict a multi word answer with a generative language model (future work \ref{cha:conclusions}).

We decided to use this kind of model knowing that our accuracy was going to be lower as we thought that our model was more innovative and more capable of being applied in \emph{real life}, meaning that we did not improved our model towards the VQA challenge or VQA datasets but to Visual Question-Answering tasks in general and to our ambitious goal (which is out of the scope of this thesis but is our future objective) of generating question-answer pairs from images. To do so, we certainly need a model able to answer with unseen answers from the training subset, to generate them. We believe that our model outperforms other participants of the challenge in flexibility and in interest from the research point of view.

\subsection{Per answer type results}
The VQA dataset annotations (answers) are classified in three different types: yes/no, number or other. Each question has assigned one of these three answer types, that allows us to better understand how our model acts given different types of questions and how good is it answering them.

Analyzing the results per answer type shown in Table \ref{tab:results} we can see a huge difference when it comes to accuracy between the yes/no answers and the number or other answer types. The latest usually need a higher comprehension of the image and the question to be able to answer them due to the type of questions (why...?, what is...?) as opposed to the more common question type \emph{is this...?} for the yes/no answer type. These difference can be better understand with the qualitative results in the following section.

%%%%%%%%%%%%%%%%%%%%%%%%%%%%%%%%%%%%%%%%%%%%%%%%%%%%%%%%%%%%%%%%%%%%%%%

\section{Qualitative results}

In this section we will present some examples of the results obtained for our best model. These results are from the validation subset, as we do not have the answers for the test subset. The following examples are grouped by accuracy, having three examples of each accuracy, one per question type (yes/no, number and other). The VQA evaluation script punctuate with 5 different accuracies (0, 30, 60, 90, 100) following equation\footnote{Even if the challenge provide the mentioned formula to compute the accuracy, it seems that the script is rounding the result of $(\# humans that said ans)/3$ to the closest lower integer} \ref{eq:accuracy}. 

These examples have been chosen randomly from the results in order to obtain a sample as representative as possible of the whole dataset.

\begin{figure}[H]
\centerline{
\includegraphics[width=1.2\textwidth]{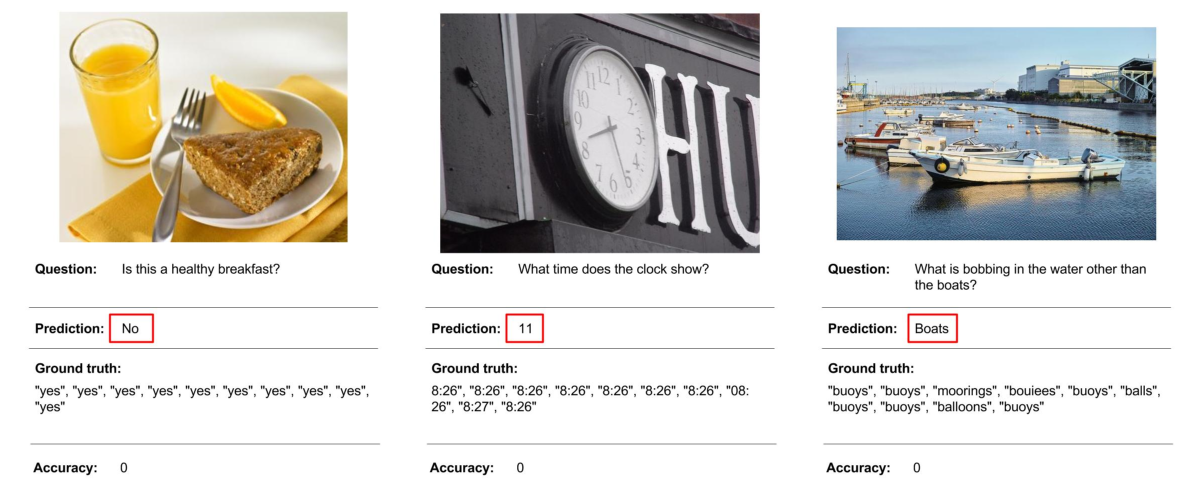}
}
\caption{Result examples with accuracy 0}
\label{fig:example-acc-0}
\end{figure}

\begin{figure}[H]
\centerline{
\includegraphics[width=1.2\textwidth]{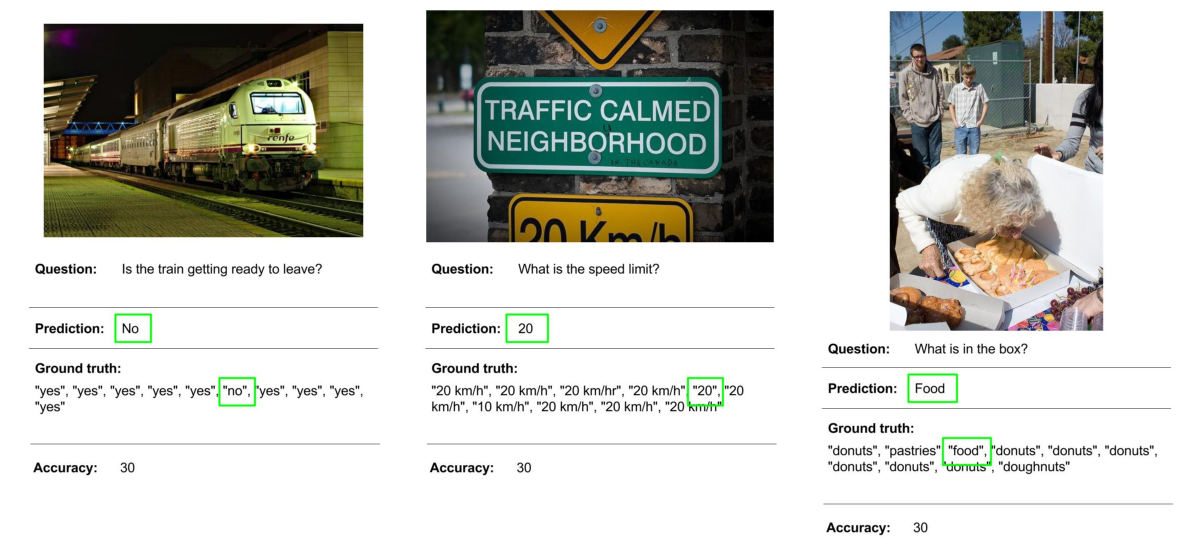}
}
\caption{Result example with accuracy 30}
\label{fig:example-acc-30}
\end{figure}

\begin{figure}[H]
\centerline{
\includegraphics[width=1\textwidth]{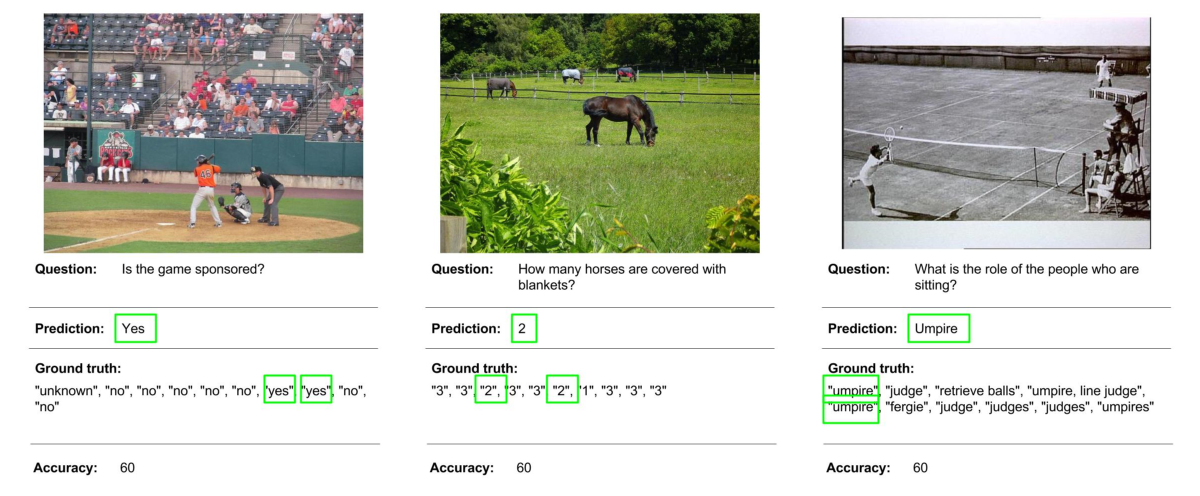}
}
\caption{Result example with accuracy 60}
\label{fig:example-acc-60}
\end{figure}

\begin{figure}[H]
\centerline{
\includegraphics[width=1\textwidth]{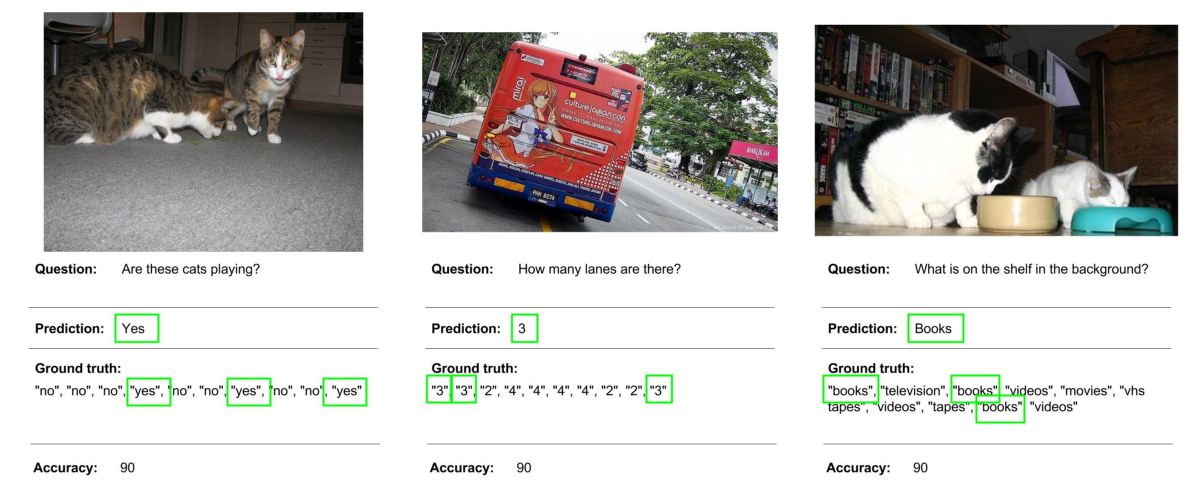}
}
\caption{Result example with accuracy 90}
\label{fig:example-acc-90}
\end{figure}

\begin{figure}[H]
\centerline{
\includegraphics[width=1\textwidth]{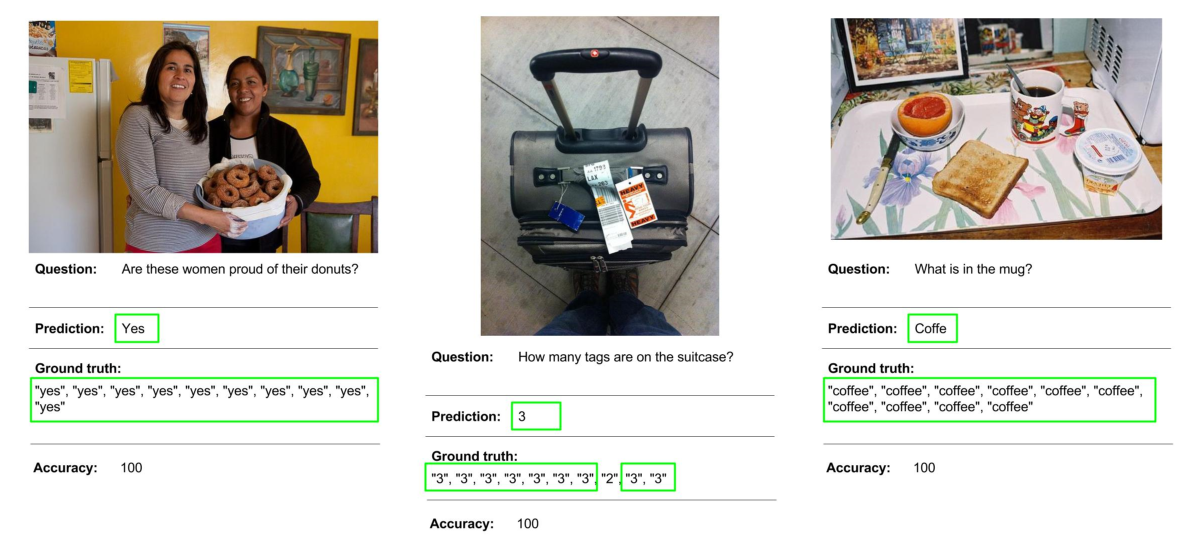}
}
\caption{Result example with accuracy 100}
\label{fig:example-acc-100}
\end{figure}

From these examples we can see that the images in the VQA dataset (that are MS COCO) are rich in information and very variate throughout the dataset. The orientation, ratio, resolution and number of channels vary from example to example, as well as the kind of content appearing.

The questions are very different in terms of what task does the model need to do in order to answer the question. These questions show perfectly the deep understanding of both the image and the question and how they are related needed to answer them. Different tasks need to be performed in order to succeed, such as sentiment analysis (Figure \ref{fig:example-acc-100}, first example), object recognition (Figure \ref{fig:example-acc-60}, second example and Figure \ref{fig:example-acc-30}, second example), Optical Character Recognition (OCR) (Figure \ref{fig:example-acc-0}, second example), activity recognition (Figure \ref{fig:example-acc-90}, first example) and so on.

As for the answers, we can appreciate why the metric provided by the challenge maybe it is not the best one to use in this task. The second example of Figure \ref{fig:example-acc-30} and the last one in Figure \ref{fig:example-acc-60} show that the predicted answer was indeed correct, but due to rephrasing and some details, only 30 and 60 of accuracy was given to them. The annotation errors also distorts the results, as in Figure \ref{fig:example-acc-60} second example, where the correct answer is 3 and, even if we predicted 2, the script evaluated our answer with 60.

%%%%%%%%%%%%%%%%%%%%%%%%%%%%%%%%%%%%%%%%%%%%%%%%%%%%%%%%%%%%%%%

\section{Some words at the VQA Challenge}
At date of 26 of June, 2016, some of our team's members assist to the Computer Vision and Pattern Recognition 2016 at Las Vegas, USA, to present our extended abstract "Towards Automatic Generation of Question Answer Pairs from Images" in the Visual Question-Answering workshop. In this workshop several speakers explained their research in the VQA field, including the winner of the VQA challenge. 

We want to highlight some remarkable comments from the last session. \\
Margaret Mitchell, from Microsoft Research, mentioned the interest of generating questions and answers from image as an extension to VQA. Mostafazadeh \emph{et. al.} (including Mitchell) have recently published a paper where they propose a model to generate natural questions \cite{mostafazadeh2016generating} (we presented our extended abstract before this paper was published). \\
Another interesting comment, coming from Trevor Darrell (UC Berkeley), was his concern about solving VQA with a closed set of answers. This is building a classifier upon the most common seen answers in the training set (which is what a lot of the participants did) as opposite of our model which generates new answers even if the model have not seen them before in training time.

These comments supports the route we have taken to accomplish this thesis.

%%%CHAPTER 5%%%%%%%%%%%%%%%%%%%%%%%%%%%%%%%%%%%%%%%%%%%%%%
\chapter{Budget}

This project is a research study and it has not been developed with a product or service in mind that could be sold in the marketplace. We have used the computational resources provided by the \emph{Grup de Processat d'Imatge} of UPC, so there has not been any additional cost in terms of hardware. 

The hardware resources needed for this project were a CPU and a GPU with at least 12GB of GPU RAM and over 50GB of regular RAM. To be able to estimate the cost of the hardware in this project we will use the Amazon Web Services (AWS) Elastic Compute Cloud (EC2) service as they offer cloud computing resources per hour and they are a common solution for this needs. The EC2 instance more similar to our specifications is the g2.8xlarge which provides 60GB of RAM and 4 GPUs with 4GB of RAM each one. The cost of this service is \$2,808 per hour which is \$67,40 per day. We spend 60 days approximately using the computing resources, thus giving an approximate cost of \$4.043,52 for the hardware needs.

Regarding software, everything we have used is open-source and thus this does not add any cost.

Being said that, the only \emph{real} cost we can deduce from this project could be the salary of the team involved in developing it.
Basically three members have formed this team: a senior engineer as the advisor, a junior engineer as the co-advisor and myself as a junior engineer.

As presented in the workplan's Gantt \ref{fig:gantt} the total duration of the project has been 24 weeks but the first weeks of work were only personal research. The other difference in the number of weeks is due to the fact that the co-advisor joined the project after few weeks.

\begin{table}[h]
\centering
\begin{tabular}{| m{10em}| m{4em} | m{5em} | m{5em} | m{5em} |} 
\hline
 & \textbf{Weeks} & \textbf{Wage/hour} & \textbf{Dedication} & \textbf{Total}  \\ [0.5ex]
\hline
Junior engineer & 24  & 10,00 \euro /h & 25 h/week & 6,000 \euro\\ [0.5ex]
\hline
Junior engineer & 16  & 10,00 \euro /h & 4 h/week & 640 \euro\\ [0.5ex]
\hline
Senior engineer & 20 & 20,00 \euro /h  & 4 h/week & 1.600 \euro\\ [0.5ex]
\hline
 \multicolumn{4}{|r|}{\textbf{Total}} & 8.240 \euro   \\ [0.5ex]
\hline
\end{tabular}
\caption{Budget of the project}
\end{table}

%%%CHAPTER 6%%%%%%%%%%%%%%%%%%%%%%%%%%%%%%%%%%%%%%%%%%%%%%
\chapter{Conclusions}
\label{cha:conclusions}

When we started this thesis we had three main goals in mind. The first one was to be able to build a model for VQA and present our results to the CVPR16 VQA Challenge. The second was to, through the process of building the VQA model, have a better understanding of the techniques used to process text in the deep learning framework. We also wanted to explore how to combine text and visual features together. Our last goal was to build the software around this project as modular, reusable and following best practices as possible.

Looking back to our results and all the work presented here, we believe that we have accomplished all three goals successfully. This has not been an easy journey and we are not saying that there is no space for improvements. As we have seen in table \ref{tab:results} there is still a big gap to fill in terms of accuracy for VQA models.

At the beginning we started the project with little knowledge of how Recurrent Neural Networks (RNN) worked and how to apply them to process sequences of text. Building our first QA model only for text gave us the needed expertise to move forward to the more complex systems that we wanted to face, VQA. A remarkable point here was the incorporation of Santiago Pascual to the team, which helped us a lot in the process of gaining this understanding of the RNN and the NLP world.

Having this QA model as a starting point, we started developing new VQA models that could merge the text and visual information, but not without reaching dead ends such as the use of VGG-16 (which, even if it is possible to use it, we could not due to timing constraints). We tried different model's configuration parameters and architectures and through this iterative process of modifying the model and checking its performance we gain this notion of how the model is affected by those parameters and also we noticed that the complexity of the task does not give much space in terms of adjusting the parameters. With that we mean that the models worked at some specific range of values which was not very large.

Finally we could train and publish a model with a similar accuracy of the baseline one defined by the VQA organizers but more prone to extending it and improving it.

We would like to highlight that during the course of this project we presented an extended abstract to the CVPR16 VQA workshop and it was accepted. This extended abstract with its own poster was presented in the VQA workshop at 26th June 2016. The extended abstract exposed one of our ideas of future work.

Having this in mind, as a future work we are planing to take this last model and attach a generative language model at the end so it can predict multiple word answers. We believe that making this improvement we will be able to outperform the baseline. \\
Another improvement that we are thinking about is to change the sentence embedding of the question to a character embedding, which exploits even more information of the words such as the relation between prefix and sufixes. \\
As we have already mention, we also want to actually implement the ideas in our extended abstract to create a model that is able to generate Question-Answer Pairs (QAP) from an image.

% %%%APPENDIX %%%%%%%%%%%%%%%%%%%%%%%%%%%%%%%%%%%%%%%%%%%%%%
 
\chapter{Appendices}
\label{cha:appendices}

In this appendices you will find our accepted extended abstract presented in the CVPR16 VQA workshop and the poster we presented in Las Vegas, Nevada (USA) on June 26, 2016.

\includepdf[pages={-}]{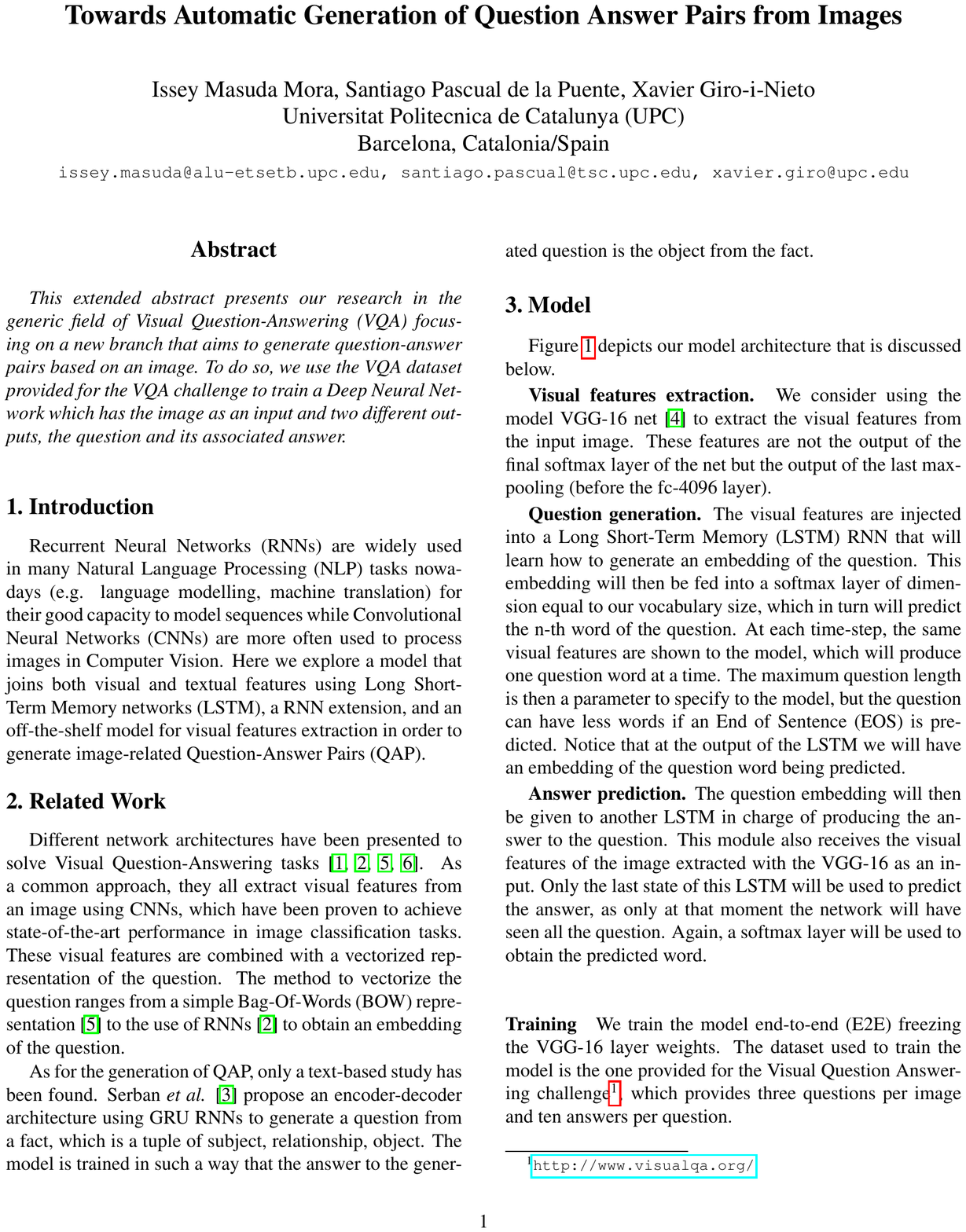}

\includepdf[pages={-},angle=90]{appendices/poster_cvpr_vqa_workshop.pdf}

%%%BIBLIOGRAPHY %%%%%%%%%%%%%%%%%%%%%%%%%%%%%%%%%%%%%%%%%%

\bibliographystyle{plain}
\bibliography{ref}

\end{document}